\documentclass[review]{elsarticle}

\usepackage{lineno,hyperref}
\modulolinenumbers[5]

\journal{Journal of \LaTeX\ Templates}

\usepackage{latexsym}
\usepackage[fleqn]{amsmath}

 \usepackage{amsfonts}
\usepackage{amsmath}
\newcommand{\argmin}{\operatornamewithlimits{argmin}}
\newcommand{\mb}{\mathbb} 
\newcommand{\vc}{\mathbf} 

\newcommand{\mc}{\mathcal}
\newcommand{\mr}{\mathrm}
\newtheorem{thm}{Theorem}[section]

\newtheorem{lem}{Lemma}[section]

\begin{document}
\begin{frontmatter}
\title{Influence Function  and Robust Variant of Kernel Canonical Correlation Analysis}

\author[mymainaddress]{Md. Ashad Alam\fnref{myfootnote}\corref{mycorrespondingauthor}}
\cortext[mycorrespondingauthor]{Corresponding author (E-mail address): malam@tulane.edu}
\author[mysecondaryaddress]{Kenji Fukumizu}
\author[mymainaddress]{Yu-Ping Wang}
\address[mymainaddress]{Department of Biomedical Engineering,   Tulane University, New Orleans, LA 70118, USA}
\address[mysecondaryaddress]{ The Institute of Statistical Mathematics, Tachikawa, Tokyo 190-8562, Japan.}
\fntext[myfootnote]{Department of Statistics, Hajee Mohammad Danesh Science and Technology\\ University Dinajpur 5200, Bangladesh.}




\begin{abstract}
Many  unsupervised kernel methods  rely  on  the estimation of the  kernel covariance operator (kernel CO) or kernel cross-covariance operator (kernel CCO). Both kernel CO and kernel CCO are sensitive to contaminated data, even when bounded positive definite kernels are used. To the best of our knowledge, there are few well-founded robust kernel methods for statistical unsupervised learning. In addition, while the influence function (IF) of an estimator can characterize its robustness,  asymptotic properties and standard error,  the IF of a standard kernel canonical  correlation analysis (standard kernel CCA) has not been derived yet. To fill this gap,  we first propose a robust kernel covariance operator (robust kernel CO) and a robust  kernel cross-covariance operator (robust kernel CCO)  based on  a generalized loss function instead of the quadratic loss function. Second, we derive the IF for robust kernel CCO and standard kernel CCA. Using the IF of the standard kernel CCA, we can detect influential observations from two sets of data. Finally, we propose a  method based on the robust kernel CO and the robust kernel CCO, called {\bf robust kernel CCA}, which is less sensitive to noise than the standard kernel CCA. The introduced principles  can also be applied  to  many other kernel methods involving  kernel CO or kernel CCO. Our experiments on synthesized data and imaging genetics analysis demonstrate that the  proposed IF of standard kernel CCA can identify outliers. It is also seen that  the proposed  robust  kernel CCA  method performs better for  ideal  and contaminated data than the standard kernel CCA.
\end{abstract}

\begin{keyword}
Robustness, Influence function, Kernel (coss-) covariance operator, Kernel methods, and Imaging genetics analysis.
\end{keyword}

\end{frontmatter}


\section{Introduction}
\label{sec:Intro}
To accelerate the analysis of  complex data, kernel based methods (i.e., the support vector machine, kernel ridge regression, multiple kernel learning, kernel dimension reduction in regression, and so on) have proved to be powerful  techniques and have been actively studied over the last two decades   due to their many flexibilities \citep{SVM92,Saunders98,Charpiat-15,  Back-08,Steinwart-08,Hofmann-08}.  Examples of  unsupervised kernel methods include kernel principal component analysis (kernel PCA), kernel canonical correlation analysis (standard kernel CCA), and  weighted multiple kernel CCA  \citep{Schlkof-kpca,Akaho,Back-02,Ashad-14,Yu-11}. These methods have been  extensively studied for decades in the use of unsupervised kernel methods. 
However,  all of these approaches  are not robust and are sensitive to the contaminated model. This paper introduces the robust kernel covariance operator (kernel CO) and kernel cross-covariance operator (kernel CCO) for unsupervised kernel methods such as kernel CCA.

 Although many researchers have been studying the robustness issue in a  supervised learning setting (e.g., the support vector  machine for classification and regression \citep{Christmann-04,Christmann-07,Debruyne-08})  there are generally few well-founded robust methods for kernel unsupervised learning.  The robustness is an important and challenging issue in using statistical machine learning for  multiple source data analysis. This is  because {\bf outliers} often occur in  real data, which can wreak havoc when used in   statistical machine learning methods. Since 1960s, many robust methods,  which are  less sensitive to outliers,  have been developed to overcome this problem.  The objective of robust statistics is to use the methods  from  the bulk of the data and detect the deviations from the original patterns \citep{Huber-09,Hampel-11}.

Recently, in the field of kernel methods,  a robust kernel density estimator (robust kernel DE) based on robust kernel mean elements (robust kernel ME) has been proposed by \cite{Kim-12}, which  is less sensitive to outliers than the kernel density estimator. Robust kernel DE is computed using a kernelized iteratively re-weighted least squares (KIRWLS) algorithm in a reproducing kernel Hilbert space (RKHS). In addition,  two spatial robust  kernel PCA methods have been  proposed based on the weighted eigenvalue decomposition  \citep{Huang-KPCA} and  spherical kernel PCA \citep{Debruyne-10}, showing that the influence function (IF) of kernel PCA, a well-known measure of robustness, can be  arbitrarily large for unbounded kernels.

The  kernel methods explicitly or implicitly depend on the kernel CO or the kernel CCO. These operators are among the most useful tools in  unsupervised kernel methods but have not yet been robustified. This paper shows that they can be formulated as an empirical optimization problem  to achieve robustness by combining empirical optimization problems with the idea of Huber or Hampel on the  M-estimation model \citep{Huber-09,Hampel-11}. The {\bf robust kernel CO} and {\bf robust kernel CCO} can be computed efficiently via a KIRWLS algorithm.

In the past decade, CCA with a positive definite kernel has been proposed and is called {\bf standard kernel CCA}. Several of its variants have also been proposed \citep{Fukumizu-SCKCCA, Hardoon2009, Otopal-12,Ashad-15}. Due to the use of simple  eigen decomposition, they are still a well-used method for multiple source data analysis.  An empirical comparison and sensitivity analysis  for robust linear CCA and standard kernel CCA have also been discussed, which give a similar interpretation as kernel PCA but without any robustness measure  (e.g., IF of standard kernel CCA) \citep{Ashad-10}.  In addition, the author in \cite{Romanazii-92} has  proposed the IF  of canonical correlation and canonical vectors of linear CCA.  While the  IF of an estimator can characterize its robustness, asymptotic properties and standard error, the IF of standard kernel CCA has not yet  been proposed. In addition, a robust kernel CCA  has   not yet been studied.  All of these considerations provide motivation  to study the IF of kernel CCA and  the robust  kernel CCA in unsupervised learning.

The contribution of this paper is fourfold. First, we propose  a robust kernel  CO and robust kernel CCO  based on a generalized loss function instead of the quadratic loss function. Second, we propose the IF of  standard kernel CCA: kernel canonical correlation (kernel CC) and kernel canonical variates (kernel CV). Third, we propose a method for detecting the influential observations from multiple sets of data, by proposing a visualization method using the IF of kernel CCA. Finally, we propose a  method based on robust kernel CO and robust kernel CCO, called {\bf robust kernel CCA}, which is less sensitive than standard kernel CCA. Experiments on both synthesized data and imaging genetics analysis demonstrate that the proposed visualization and  robust  kernel CCA can be applied effectively to  ideal and contaminated data.

The remainder of this paper is organized as follows. In the following  section, we provide a brief review of positive  definite kernel,  kernel  ME, robust kernel ME and  kernel  CCO.  In Section \ref{sec:RKCCO} we  present the definition, representer theorem,  KIRWLS convergence,  and a  algorithm of  robust kernel CCO. In Section \ref{sec:if},  we discuss the basic notion of the IF, the IF of kernel ME,  kernel CO,  kernel CCO and robust kernel CCO. After a brief review of standard kernel CCA in  Section \ref{sec:CKCCA}, we propose the  IF of standard  kernel CCA (kernel CC and  kernel CV) and the robust kernel CCA  in Section \ref{sec:IFKCCA}  and in Section \ref{sec:RKCCA}, respectively. In Section \ref{sec:Exp}, we describe  experiments conducted on both synthesized data and  real imaging genetics analysis.  In Section \ref{sec:conld}, concluding remarks and future research directions are presented.  In the appendix, we discuss the detailed results.

\section{Standard and robust kernel  (cross-) covariance operator}
The kernel ME, kernel CO, and kernel CCO  with positive definite kernel have been extensively applied to nonparametric statistical inference  through representing distributions in the form of means and covariance in the RKHS \citep{Gretton-08, Fukumizu-08,Song-08, Kim-12,Gretton-12}. To define  the kernel ME, robust kernel ME, kernel CO and kernel CCO,  we need the basic notions of positive definite kernels and Reproducing kernel Hilbert space (RKHS), which are briefly addressed in the following \citep{Aron-RKHS, Berlinet-04, Ashad-14T}.

\subsection{Basic notion of kernel methods}
\label{sec:basic}
Let $F_X$, $F_Y$  and $F_{XY}$ be  probability measures on the given nonempty sets $\mc{X}$, $\mc{Y}$ and  $ \mc{X}\times\mc{Y}$, respectively,  such that $F_X$ and $F_Y$ are the marginals of $F_{XY}$.  Also let $X_1, X_2, \ldots, X_n$;  $ Y_1, Y_2, \ldots, Y_n$ and $(X_1, Y_1),(X_2, Y_2), \ldots, (X_n, Y_n)$ be the independent and identically distributed (IID)  samples from the distribution  $F_X$, $F_Y$  and $F_{XY}$, respectively. A symmetric kernel, $k(\cdot,\cdot)\colon \mc{X}\times \mc{X} \rightarrow \mb{R}$, defined on a space  is called a {\bf positive definite kernel} if the Gram matrix $(k(X_i, X_j))_{ij}$ is positive semi-definite for all $i,j \in \{1, 2, \cdots, n\}$. A RKHS is a Hilbert space with a reproducing kernel whose span is dense in the Hilbert space. We can equivalently define an RKHS as a Hilbert space of functions with all evaluation functionals bounded and linear. The Moore-Aronszajn theorem states that every symmetric, positive definite kernel defines a unique reproducing kernel Hilbert space \cite{Aron-RKHS}. The {\bf feature map} is a  mapping $ \vc{\Phi}: x \to \mc{H}_X$ and  defined as $\vc{\Phi}(\cdot) = k(\cdot, x), \forall\, x\in \mc{X}$). The vector $\vc{\Phi}(x) \in \mc{H}_X$ is called  a {\bf feature vector}. The inner product of two feature
vectors can be defined as  $\langle  \vc{\Phi} (x), \vc{\Phi} (x^\prime)\rangle_{\mc{H}_X} = k(x, x^\prime)$ for all $x, x^\prime \in \mc{X}$. This is called  the {\bf kernel trick}. By the reproducing property, $f(x)=\langle f(\cdot), k(\cdot, x)\rangle_{\mc{H}_X}$, with   $f \in \mc{H}_X$  and the kernel trick, the kernel can evaluate the inner product of any two feature vectors efficiently, without knowing an explicit form of either the feature map or the feature vector. Another great advantage is that the computational cost does not depend on the dimension of the original space after computing the Gram matrices \citep{Fukumizu-14,Ashad-14}.

\subsection{Standard kernel mean element}
\label{sec:me}
Let $k_X$ be a measurable positive definite kernel on $\mc{X}$ with $\mb{E}_X[\sqrt{k(X, X)}] < \infty$.  The {\bf kernel mean}, $\mc{M}_X$, of $X$ on  $\mc{H}_X$ is an element of $\mc{H}_X$ and is defined by the mean of the $\mc{H}_X$-valued random  variable $k_X(\cdot, X)$,  
\[ \mc{M}_X(\cdot)=\mb{E}_X[k_X(\cdot, X)].\]
The kernel mean always exists with arbitrary probability  under the assumption that positive definite kernels are bounded and measurable.
By the reproducing property,  the kernel ME satisfies the following equality
 \[\langle \mc{M}_X, f \rangle_{\mc{H}_X} = \langle \mb{E}_X[k_X(\cdot, X)], f \rangle_{\mc{H}_X} = \mb{E}_X\langle  k_X(\cdot, X), f \rangle_{\mc{H}_X}=\mb{E}_X[f(X)],\] for all    $f\in \mc{H}_X$.

 The  empirical kernel ME, $\widehat{\mc{M}}_X=\frac{1}{n}\sum_{i=1}^n\vc{\Phi}(X_i)=  \frac{1}{n} \sum_{i=1}^n k_X(\cdot, X_i)$ is an  element of the RKHS,
\[\langle \widehat{\mc{M}}_X, f \rangle_{\mc{H}_X}= \langle  \frac{1}{n}\sum_{i=1}^n k_X(\cdot, X_i), f\rangle =  \frac{1}{n}\sum_{i=1}^n  f(X_i).\]
The  empirical kernel ME of  the feature vectors $\vc{\Phi}(X_i)$  can be regarded as a solution to the empirical risk optimization problem \citep{Kim-12}
\begin{eqnarray}
\label{EROP1}
\widehat{\mc{M}}_X=\argmin_{f\in \mc{H}_X} \sum_{i=1}^n\| \vc{\Phi}(X_i)- f\|^2_{\mc{H}_X}.
\end{eqnarray}

\subsection{Robust kernel mean element}
\label{sec:rkme}
As explained in  Section \ref{sec:me},  the   kernel ME is the  solution to the empirical risk optimization problem, which is a least square type of  estimator.  This type of estimator is sensitive  to the presence of  outliers in the feature, $\vc{\Phi}(X)$.  To reduce the effect of outliers, we can use $M$-estimation. In recent years, the robust kernel ME has been proposed for density estimation \citep{Kim-12}. The robust kernel ME,  based on  a robust loss function  $\zeta(t)$ on $t \geq 0$, is defined as
\begin{eqnarray}
\label{EKME1}
\widehat{\mc{M}}_R=\argmin_{f\in \mc{H}_X} \sum_{i=1}^n\zeta(\| \vc{\Phi}(X_i)- f\|_{\mc{H}_X}).
\end{eqnarray}
Examples of robust loss functions include  Huber's loss function,  Hampel's loss function, or Tukey's  biweight loss function. Unlike the quadratic loss function, the derivative of these loss functions  are bounded \citep{Huber-09, Hampel-86, Tukey-77}. The  Huber's function, a hybrid approach between squared and absolute  error losses, is defined as:
\begin{eqnarray}
\zeta(t)=
\begin{cases}
	 t^2/2,\qquad  \qquad 0\leq t\leq c
\\
 ct-c^2/2,\qquad c\leq t, \nonumber
\end{cases}
\end{eqnarray}
where c ($c >0$) is a tuning parameter. The Hampel's loss function is defined as:
\begin{eqnarray}
\zeta(t)=
\begin{cases}
	 t^2/2,\qquad  \qquad\qquad\qquad 0\leq t\le c_1
\\
 c_1t-c_1^2/2,\qquad\qquad  c_1\leq t < c_2
\\
-\frac{c_1}{2(c_3-c_2)}(t-c_3)^2+ \frac{c_1(c_2+c_3-c_1)}{2},\qquad  c_2\leq t < c_3\\
 \frac{c_1(c_2+c_3-c_1)}{2}, \qquad\qquad  c_3\leq t, \nonumber
\end{cases}
\end{eqnarray}
where the non-negative free parameters $c_1 < c_2 < c_3$ allow us to control the degree of suppression of large errors. The Tukey's biweight loss functions is defined as:
\begin{eqnarray}
\zeta(t)=
\begin{cases}
	 1-(1-(t/c)^2)^3,\qquad  \qquad 0\leq t\leq c
\\
 1,\qquad c\leq t, \nonumber
\end{cases}
\end{eqnarray}
where $c >0$.

\begin{figure}[t]
\begin{center}
\includegraphics[width=\columnwidth, height=8cm]{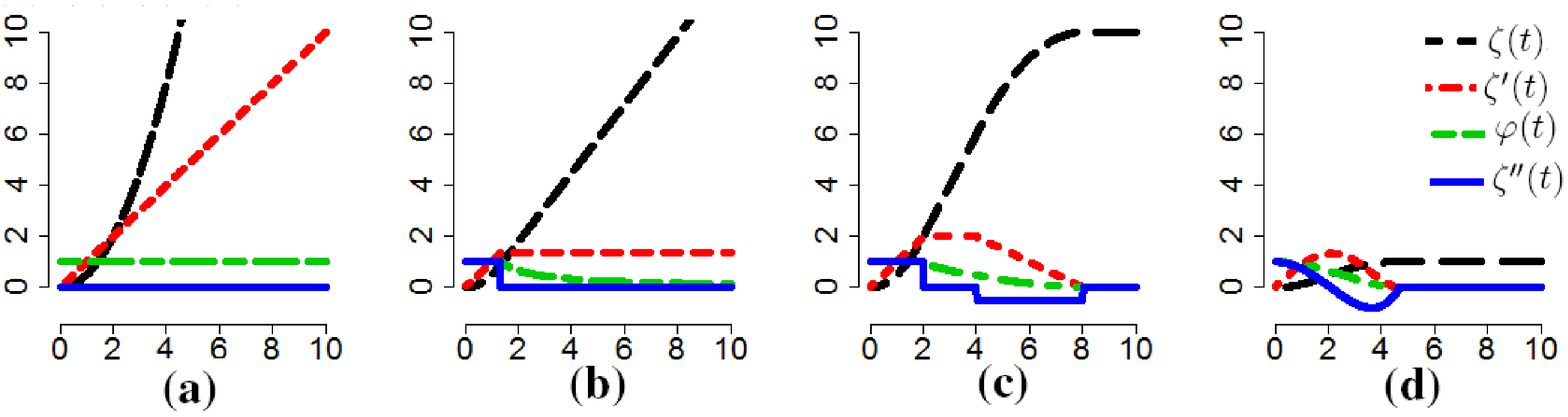}
\caption{Family of  loss functions (a) Quadratic loss (b) Huber's loss (c) Hampel's loss  and (d) Tukey's loss.}
\label{Me}
\end{center}
\end{figure}

The basic assumptions of the loss functions are; (i) $\zeta$ is non-decreasing, $\zeta(0)=0$ and  $\zeta(t)/t \to 0$ as $t\to 0$, (ii) $\varphi(t)=\frac{\zeta^\prime(t)}{t}$  exists and is finite, where $\zeta^\prime(t)$ is the derivative of $\zeta(t)$, (iii) $\zeta^\prime(t)$ and $\varphi(t)$ are continuous, and bounded, and  (iv) $\varphi(t)$ is Lipschitz  continuous. All of these assumptions hold for Huber's loss function as well as others \citep{Kim-12}.  Figure  \ref{Me}  presents the family of  loss functions, $\zeta (t)$,  $\zeta^\prime(t)$,  $\varphi(t)$, and $\zeta^{\prime\prime}(t)$ (second derivative of $\zeta (t)$).

Essentially  Eq. (\ref{EKME1})  does not have a closed form solution, but using KIRWLS, the solution of robust kernel mean  is,
\[\widehat{\mc{M}}_R^{(h)}= \sum_{i=1}^n w_i^{(h-1)}k_X(\cdot, X_i),\]
where $w_i^{(h)}=\frac{\varphi(\|\tilde{\vc{\Phi}}(X_i) - f^{(h)}\|_{\mc{H}_X})}{\sum_{b=1}^n\varphi(\| \vc{\Phi}(X_b)- f^{(h)}\|_{\mc{H}_X})}\,, \rm{and} \, \varphi(x)=\frac{\zeta^\prime(x)}{x}.$

Given the weights of the robust kernel ME, $\vc{w}=[ w_1, w_2, \cdots, w_n]^T$, of  a set of observations $X_i, \cdots, X_n$,  the points $\tilde{\vc{\Phi}}(X_i):= \vc{\Phi}(X_i) - \sum_{a=1}^nw_a \vc{\Phi}(X_a)$  are centered and the centered robust Gram matrix is $\tilde{K}_{ij}= \langle \tilde{\vc{\Phi}}(X_i),\tilde{\vc{\Phi}}(X_j)\rangle= (\vc{C}\vc{K}_X\vc{C}^T)_{ij}$,
where  $\vc{K}_X = (k_X (X_i, X_j))_{i=1}^n$ is a  Gram matrix, $\vc{1}_n=[1_1, 1_2, \cdots, 1_n]^T$ and $\vc{C}=\vc{I}- \vc{1}_n\vc{w}^T$.
For a set of test points $ X^t_1, X_2^t, \cdots, X_T^t$,  we define two matrices of order $T\times n$ as  $K_{ij}^{test}= \langle \vc{\Phi}(X^t_i), \vc{\Phi}(X_j) \rangle$ and  $\tilde{K}_{ij}^{test}= \langle \vc{\Phi}(X^t_i)- \sum_{b=1}^nw_b \vc{\Phi}(X_b), \vc{\Phi}(X_j)- \sum_{d=1}^nw_d  \vc{\Phi}(X_b)\rangle$.  Like the centered Gram matrix,  the  centered robust Gram matrix of test points,  $K_{ij}^{test}$,  in terms of  the robust Gram matrix and $\vc{1}_t=[1_1, 1_2, \cdots, 1_t]^T$  is defined as,
\begin{eqnarray}
\label{CM2}
\tilde{K}_{ij}^{test} =  (\vc{K}^{test}- \vc{1}_t\vc{w}^T \vc{K} - \vc{K}^{test} \vc{w}\vc{1}_n^T + \vc{1}_t\vc{w}^T \vc{K} \vc{w}\vc{1}_n^T)_{ij} \nonumber
\end{eqnarray}
\subsection{Standard kernel (cross-) covariance operator}
\label{sec:kernel CCO}
In this section we study the covariance of two random feature vectors $k_X(\cdot, X)$ and $k_Y(\cdot, Y)$.  As for the standard random vectors,  the notion of kernel covariance is useful as the basis in describing the statistical dependence among two or more variables.

Let $(\mc{X}, \mc{B}_X)$ and  $(\mc{Y}, \mc{B}_Y)$ be  two measurable spaces and $(X, Y)$ be a  random variable on $\mc{X}\times\mc{Y}$ with distribution $F_{XY}$. The  kernel  CCO (centered)  is a linear  operator $\Sigma_{XY} := \mc{H}_Y \to \mc{H}_X$ defined as
\[ \Sigma_{XY} = E_{XY}[\tilde{\vc{\Phi}}(X) \otimes \tilde{\vc{\Phi}}(Y)],\]
where $\tilde{\vc{\Phi}}(\cdot)=\vc{\Phi}(\cdot)- E[\vc{\Phi}(\cdot)]$ and  $\otimes$ is a tensor product operator $((a_1\otimes b_1)x= \langle x, b_1\rangle a_1\, \rm{and}\,\langle a_1\otimes a_2, b_1\otimes b_2 \rangle_{\mc{H}_{12}} = \langle a_1, b_1 \rangle_{\mc{H}_{1}}  \langle a_2, b_2 \rangle_{\mc{H}_{2}},\, \forall \, a_1, b_1 \in \mc{H}_1,\, \rm{and}\, a_2, b_2 \in \mc{H}_2$, where $\mc{H}_1$ and $\mc{H}_2$ are Hilbert spaces) \citep{Reed-80}.

Given two $k_X$ and $k_Y$ measurable positive definite kernels with respective RKHS $\mc{H}_X$ and $\mc{H}_Y$. By the reproducing property, the   kernel CCO,  with $\mb{E}_X[k_X(X, X)] < \infty$, and $\mb{E}_Y[k_Y(Y, Y)] < \infty$ is satisfied
\begin{eqnarray}
\langle f_X,  \Sigma_{XY}f_Y\rangle_{\mc{H}_X} &=&\langle f_X,  E_{XY}[\tilde{\vc{\Phi}}(X) \otimes \tilde{\vc{\Phi}}(Y)]f_Y\rangle_{\mc{H}_X} \nonumber \\
 &=& \mb{E}_{XY}\left[\langle f_X, k_X(\cdot, X) - \mc{M}_X \rangle_{\mc{H}_X} \langle f_Y, k_Y(\cdot, Y) - \mc{M}_Y \rangle_{\mc{H}_Y} \right]\nonumber\\&=& \mb{E}_{XY}\left[(f_X(X) - E_X[f(X)]) (f_Y(Y) - E_Y[f(Y)])\right] \nonumber
\end{eqnarray}
for all $f_X\in \mc{H}_X$ and $f_Y\in \mc{H}_Y$. This is a bounded operator.  As  shown in   Eq. (\ref{EROP1}), we can define kernel CCO  as  an empirical risk optimization problem as follows,
\begin{eqnarray}
\label{EROP2}
\hat{\Sigma}_{XY}=\argmin_{\Sigma\in \mc{H}_X\otimes \mc{H}_Y}\sum_{i=1}^n \|\tilde{\vc{\Phi}}(X_i) \otimes \tilde{\vc{\Phi}}(Y_i) - \Sigma\|^2_{\mc{H}_X\otimes\mc{H}_Y}.
\end{eqnarray}
The empirical kernel CCO is then
\begin{eqnarray}
\hat{\Sigma}_{XY} &=& \frac{1}{n}\sum_{i=1}^n\left(k_X(\cdot, X_i)- \frac{1}{n}\sum_{b=1}^nk_X(\cdot, X_b)\right)\otimes \left(k_Y(\cdot, Y_d)- \frac{1}{n}\sum_{d=1}^n k_Y(\cdot, Y_d) \right) \nonumber\\
 &=& \frac{1}{n}\sum_{i=1}^n \tilde{k}_X(\cdot, X_i)\otimes\tilde{k}_Y(\cdot, Y_i),
\end{eqnarray}
where  $\tilde{k}_X$ and $\tilde{k}_Y$ are centered kernels. For the special case,  when $Y$ is equal to $X$, it  gives a kernel CO.

\section{Robust kernel (cross-) covariance operator}
\label{sec:RKCCO}
Because a  robust kernel ME (see Section \ref{sec:rkme}) is used, to reduce the effect of outliers, we propose to use  $M$-estimation  to find a robust sample covariance  of  $\vc{\Phi}(X)$ and $\vc{\Phi}(Y)$.  To do this, we  estimate kernel CO and kernel CCO   based on robust loss functions, namely,  robust kernel CO and   robust kernel CCO, respectively.  Eq. (\ref{EROP2}) can be written as
\begin{eqnarray}
\label{REROP1}
\widehat{\Sigma}_{RXY}= \argmin_{\Sigma\in {\mc{H}_X\otimes\mc{H}_Y}} \sum_{i=1}^n \zeta(\| \tilde{\vc{\Phi}}(X_i) \otimes \tilde{\vc{\Phi}}(Y_i) - \Sigma\|_ {\mc{H}_X\otimes\mc{H}_Y}).
\end{eqnarray}

\subsection{Representation of  robust kernel (cross-) covariance operator}
In this section, we represent $\widehat{\Sigma}_{RXY}$  as a weighted combination of the  product of two kernels $k_X(\cdot, X_i)k_Y(\cdot, Y_i)$. We will also address necessary and sufficient conditions for the robust kernel CCO. Eq (\ref{REROP1}) can be reformulated as
$\hat{\Sigma}_{RXY} = \argmin_{\Sigma\in {\mc{H}_X\otimes\mc{H}_Y}}
J(\Sigma),$ where

\begin{eqnarray}
\label{RPe1}
J(\Sigma)=\sum_{i=1}^n \zeta(\| \tilde{\vc{\Phi}}(X_i) \otimes \tilde{\vc{\Phi}}(Y_i) - \Sigma\|_ {\mc{H}_X\otimes\mc{H}_Y}).
\end{eqnarray}
In order to optimize $J$ in a product RKHS, the necessary conditions are characterized through the  G\^{a}teaux differentials of $J$. Given a product vector space $\mc{X}\times\mc{Y}$ and a function $A : \mc{X}\times\mc{Y} \to [-\infty, \infty]$, the G\^{a}teaux differential of $A$ at $Z=(X,Y)\in \mc{X}\times\mc{Y}$ with incremental $\mc{T} \in \mc{X}\times\mc{Y}$  is defined as
\[\partial A (Z; \mc{T})= \lim_{\epsilon\to 0} \frac{A(Z+\epsilon\mc{T})+ A(Z)} {\epsilon}.\]
The  G\^{a}teaux differential on a probability distribution is  also defined in Section \ref{sec:if}.
 
  Based on the optimality principle \citep{Luenberger-97},  the G\^{a}teaux differential is well defined for all $\mc{T}$ and  a necessary condition for $A$ to have a minimum at $Z_0= (X_0, Y_0)\in \mc{X}\times\mc{Y}$ is that $\partial A (Z_0; \mc{T})=0$. We can state the following lemma.

\begin{lem}
\label{lemma1}
Under the assumptions (i) and (ii) the G\^{a}teaux differential of the objective function $J$ at $\Sigma\in \mc{H}_X\otimes \mc{H}_Y$ and incremental $\mc{T} \in \mc{H}_X\otimes \mc{H}_Y$ is
\[\delta J(\Sigma, \mc{T})=-\langle S(\Sigma), \mc{T}\rangle_{ \mc{H}_X\otimes \mc{H}_Y},\]
where $S: \mc{H}_X\otimes \mc{H}_Y \to \mc{H}_X\otimes \mc{H}_Y$ is defined as
\[S(\Sigma)=\sum_{i=1}^n\varphi(\|\tilde{\vc{\Phi}}(X_i) \otimes\tilde{\vc{\Phi}}(Y_i)-\Sigma\|_{ \mc{H}_X\otimes \mc{H}_Y})\cdot (\tilde{\vc{\Phi}}(X_i) \otimes\tilde{\vc{\Phi}}(Y_i)-\Sigma).\]
A necessary condition for $\Sigma=\widehat{\Sigma_{RXY}}$, robust kernel CCO   is  $S(\Sigma)=\vc{0}.$
 \end{lem}
 The key difference of Lemma \ref{lemma1} and  Lemma $1$ of \cite{Kim-12} is the RKHS. The latter lemma is based on a single RKHS $\mc{H}_X$ but the former one is on a product RKHS $\mc{H}_X\times\mc{H}_Y$.  This is a generalization  result.

\begin{thm}
\label{thm1}
Under the same assumption of Lemma \ref{lemma1}, the robust kernel CCO (centered) for any $(X, Y)\in \mc{X}\times \mc{Y}$ is then	
\begin{eqnarray}
\label{RPe2}
\widehat{\Sigma}_{RXY}(X,Y)= \sum_{i=1}^n w_i\tilde{k}(X, X_i)\tilde{k}(Y, Y_i)
\end{eqnarray}
where $w_i \geq 0$,  and  $\sum_{i=1}^n w_i=1$. Furthermore,
\begin{eqnarray}
\label{RPe3}
w_i \propto \varphi(\|\tilde{\vc{\Phi}}(X_i) \otimes\tilde{\vc{\Phi}}(Y_i)-\widehat{\Sigma}_{RXY}\|_{ \mc{H}_X\otimes \mc{H}_Y}).
\end{eqnarray}
\end{thm}
Representer Theorem \ref{thm1} tells  us that in the robust loss function, when $\varphi$ is decreasing the large value of  $\|\tilde{\vc{\Phi}}(X_i)\otimes \tilde{\vc{\Phi}}(Y_i) - \widehat{\Sigma}_{RXY}\|_{\mc{H}_X\otimes\mc{H}_Y}$,  $w_i$ will be small. Therefore, the robust kernel CCO is robust in the sense that it down-weights outlying points.

In order to state the sufficient condition for $\widehat{\Sigma}_{RXY}$ to be the minimizer of Eq. (\ref{REROP1}), we need an additional assumption on $J$.
\begin{thm}
\label{thm2}
Under the  assumptions (i), (ii), and $J$ is  strictly convex, Eq. (\ref{RPe2}), Eq. (\ref{RPe3}) and $\sum_{i=1}^nw_i=1$ are sufficient conditions for the robust kernel CCO to be the minimizer of Eq. (\ref{REROP1}).
\end{thm}
For a positive definite kernel,  $J$ becomes strictly convex for the Huber loss function.

\subsection{Algorithm for  robust kernel (cross-) covariance operator}
As explained in \citep{Kim-12}, Eq. (\ref{REROP1}) does not have a closed form solution, but  using the kernel trick the standard IRWLS can be extended to a RKHS. The solution at  {\it h}th iteration is then,
\[\Sigma^{(h)}= \sum_{i=1}^n w_i^{(h-1)}\tilde{\vc{\Phi}}(X_i)\otimes\tilde{\vc{\Phi}}(Y_i),\]
where $w_i^{(h)}=\frac{\varphi(\|\tilde{\vc{\Phi}}(X_i)\otimes \tilde{\vc{\Phi}}(Y_i) - \Sigma^{(h)}\|_{\mc{H}_X\otimes\mc{H}_Y})}{\sum_{b=1}^n\varphi(\| \tilde{\vc{\Phi}}(X_b)\otimes\tilde{\vc{\Phi}}(Y_b)- \Sigma^{(h)}\|_{\mc{H}_X\otimes\mc{H}_Y})}\,, \rm{and} \, \varphi(x)=\frac{\zeta^\prime(x)}{x}.$

\begin{thm}
\label{thm3}
Under the  assumptions (i) - (iii) and  $\varphi(t)$ is  non-increasing.  Let 
\[ U = \{ \Sigma \in \mc{H}_X\otimes\mc{H}_Y | S(\Sigma)= \vc{0} \} \]
and $\{\Sigma^{(h)} \}_{h=1}^\infty$  be the sequence produced by the KIRWLS algorithm. Then $J(\Sigma^{(h)})$  decreases monotonically  at every iteration and converges. 
\[\| \Sigma^{(h)}- U\|_{  \mc{H}_X\otimes\mc{H}_Y } \doteq  \inf_{ \Sigma \mc{H}_X\otimes\mc{H}_Y}\| \Sigma^{(h)} -\Sigma\|_{\mc{H}_X\otimes\mc{H}_Y} \to 0  \] 
as $h\to \infty$.
\end{thm}

 Theorem \ref{thm3} sates that  $\Sigma^{(h)}$ becomes  close to the set of stationary points of $J$ by increasing the number of iterations. Under the assumptions of  Theorem \ref{thm3} and  for a strictly convex set $J$, it is also granted that the   $\{\Sigma^{(h)} \}_{h=1}^\infty$ converges to $\Sigma_{RXY}$  in the  Hilbert-Schmidt norm and supremum norm.

 The algorithm for estimating  robust kernel CCO  is given in Figure \ref{Robust.cross.covariance.M}.  The input of this algorithm is a robust kernel ME.   The computational complexity of a robust kernel ME is  $\mc{O}(n^2)$ in each iteration,  where $n$ is the number of data points. The algorithm that we have presented involves finding the robust kernel  CCO  with the  dimension  $n \times n$.   A naive implementation  of the algorithm in  Figure  \ref{Robust.cross.covariance.M} would  show that  both time and memory complexity are similar to $\mc{O}(n^3)$  in each iteration. In practice, the required number of iterations is around   $50$.  A computational complexity with cubic growth in the number of data points would be a serious liability in application to large dataset.  We are able to reduce the time complexity using the low-rank approximation of the Gram matrix \citep{Drineas-05}. We can also use the  random features approach.   Random Features provide a finite-dimensional alternative to the kernel trick by instead mapping the data to an equivalent randomized feature space  \citep{Rahimi-07}.
\begin{figure}
\begin{em}
\noindent\fbox{%
    \parbox{\textwidth}{%
       Input: $D=\{(\vc{X}_1,\vc{Y}_1),  (\vc{X}_2, \vc{Y}_2), \ldots (\vc{X}_n, \vc{Y}_n) \}$. The robust centered kernel  matrix $\tilde{\vc{K}}_X$ and $\tilde{\vc{K}}_Y$  with kernel $k_X$ and $k_Y$, $\tilde{\vc{K}}_{Xi}$ and,   $\tilde{\vc{K}}_{Yi}$   are the $i$-th column of  $\tilde{\vc{K}}_X$ and $\tilde{\vc{K}}_Y$, respectively. Also  define $\vc{B}_i=\tilde{\vc{K}}_{\vc{X}_i}\otimes \tilde{\vc{K}}_{\vc{Y}_i}$,  the tensor product of two vectors. Threshold $TH$ (e.g., $10^{-8}$).
\begin{enumerate}
\item[] Set $h=1$,  $ w_i^{(0)}=\frac{1}{n}$ and $\vc{e}^{(0)} =(\rm{diag}(\tilde{\vc{K}}_X\tilde{\vc{K}}_Y) - 2[\vc{w}^{(0)}]^T \tilde{\vc{K}}_X \tilde{\vc{K}}_X+  [\vc{w}^{(0)}]^T \tilde{\vc{K}}_X\tilde{\vc{K}}_Y[\vc{w}^{(0)}]^T\vc{1}_n )^{\frac{1}{2}}.$
\item[] Do the following steps until $ \frac{|J(\Sigma_{RXY}^{(h+1)})- J(\Sigma_{RXY}^{(h)})|}{J(\Sigma_{RXY}^{(h)})} < TH,$
\begin{itemize}
\item[(1)] Solve  $w_i^{(h)}=\frac{ \varphi(e_i^{(h)})}{\sum_i^n\varphi(e_i^{(h)})}$ and make a vector $\vc{w}$ for  $i=1, 2, \cdots n$.
\item[(2)]  Calculate a $n^2\times 1$ vector,
$\vc{v}^{(h)}= \vc{B} \vc{w}^{(h)}$ and make a $n\times n$ matrix $\vc{V}^{(h)}$, where $\vc{B}$ is $n^2\times n$ matrix  that $i$-th column consists of all elements of the $n\times n$ matrix  $\vc{B}_i$.
\item[(3)]  Update the robust covariance,  $\hat{\Sigma}_{RXY}^{(h+1)}= \sum_i^nw_i^{(h)}\vc{B}_i= \vc{V}^{(h)}$.
\item[(4)] Update error,  $\vc{e}^{(h+1)} =(\rm{diag}(\tilde{\vc{K}}_X\tilde{\vc{K}}_Y)- 2[\vc{w}^{(h)}]^T \tilde{\vc{K}}_X \tilde{\vc{K}}_X+ [\vc{w}^{(h)}]^T \tilde{\vc{K}}_X\tilde{\vc{K}}_Y[\vc{w}^{(h)}]^T\vc{1}_n )^{\frac{1}{2}} $.
\end{itemize}
\item[]  Update $h$ as $h+1$.
\end{enumerate}
Output:  the robust cross-covariance operator.
    }%
}
\caption{ The algorithm for estimating robust kernel cross-covariance operator.}
\label{Robust.cross.covariance.M}
\end{em}
\end{figure}

\section{Influence function of robust kernel and kernel (cross-) covariance operator}
\label{sec:if}
To define the robustness in statistics, different approaches have been proposed,   for example, the {\bf minimax approach} \citep{Huber-64}, the {\bf sensitivity curve} \citep{Tukey-77}, the {\bf IF} \citep{Hampel-74,Hampel-86} and  the finite sample {\bf breakdown point} \citep{Dono-83}. Due to its simplicity, the IF  is the most useful approach in  statistical supervised learning \citep{Christmann-07,Christmann-04}. In this section, we briefly introduce the definition of IF and  the IF of kernel ME,  kernel CO, and  kernel CCO. We then propose the IF of robust kernel CO and the robust kernel CCO.

Let $X_1, X_2, \cdots, X_n \in \mc{X}$  is a IID  sample from a population with distribution function $F$, its empirical distribution function is $F_n$, and $T_n =  T_n( X_1, X_2, \cdots, X_n)$ is a statistic.  Also let  $\mc{A}(\mc{X})$ be  a   class of all possible distributions containing $F_n$ for all $n\geq 1$ and $F$. We assume that there exists a functional $T: \mc{D} \to \mb{R}$, where $ \mc{D}$  is the set of all probability distributions in  $\mc{A}(\mc{X})$ for which $T$ is defined, such that 
\[ T_n = T(F_n),\]
where $T$  does not depend on $n$. $T$ is then called a statistical functional. 
If  the domain of  $T$  is  a convex set containing all distributions, $\mc{D}$ and  the data do not follow the model $F$ in $\mc{D}$ exactly but  slightly  going toward a distribution $G$. The   G\^{a}teaux derivative, $T_F^\prime$ of $T$ at $F$ is defined as
\begin{eqnarray}
T_F^\prime (G-F) = \lim_{\epsilon\to 0 }\frac{T((1-\epsilon)F+\epsilon G) - T(F)}{\epsilon}.  \nonumber
\end{eqnarray}
The G\^{a}teaux differentiability at $F$ ensures the directional derivative of $T$ exists in all directions that stay in $\mc{D}$.

 Suppose $X^\prime\in \mc{X}$ and $G=\Delta_{X^\prime} \in \mc{D}$ is the probability measure which  gives mass $1$  at the point $\{X^\prime\}$. Then, $F^{\epsilon}=(1-\epsilon)F +\epsilon\Delta_{X^\prime}$ is a  $\epsilon-$ contaminated distribution.  The {\bf influence function} (special case of G\^{a}teaux Derivative) of $T$ at $F$ is defined by
\begin{eqnarray}
IF(X^\prime, T, F)=\lim_{\epsilon\to 0 }\frac{T(F^{\epsilon}) - T(F)}{\epsilon}
\end{eqnarray}
provided that the limit exists. It can be intuitively interpreted as a suitably normalized asymptotic influence of outliers on the value of an estimate or test statistic. The IF exists with an even weaker condition than G\^{a}teaux differentiability. The IF reflects the bias caused by adding a few outliers at the point $X^\prime$, standardized by the amount of contamination. Therefore a bounded IF accelerates the robustness of an  estimator \citep{Hampel-86}.

\subsection{Influence function based robustness measures}
The three metrics of the IF function that can be used for robustness measures of the functional $T$ are the gross error sensitivity, local shift sensitivity and rejection point. The gross error sensitivity of $T$ at $F$ is defined as
\begin{eqnarray}
\gamma^*=sup_{X\in {X}}|IF(X, F, R)|.
\end{eqnarray} 
The gross error sensitivity measures the worst effect that a small amount of contamination of fixed size can have on the estimator.  The local shift sensitivity of $T$ at $F$  for all $X_1, X_2\in \mc{X}$ is defined  by
\begin{eqnarray}
\lambda^{*}=sup_{X_1\not=X_2}\frac{|IF(X_1,F,T)-IF(X_2;F,T)|}{|X_1 - X_2|}.\nonumber
\end{eqnarray}
$\lambda^{*}$ measures the worst effect of rounding error (small function in the observation). The rejection point of $T$  at $F$ is defined by
\begin{eqnarray}
\rho_{\mc{F}}^*=inf \{ m >0; IF (X, F, T)=0, \, \rm{when} \, |X| > m\}. \nonumber
\end{eqnarray}
 The $\rho_{\mc{F}}^*$ is infinite if there exits no such $m$. We can reject those observations, which are farther away than $\rho_{\mc{F}}^*$.   For a robust estimator,  $\rho_{\mc{F}}^*$ will be finite.

\subsection{Influence function  of  kernel (cross-) covariance operator}
\label{sec:ifK}
In kernel methods, every estimate is a function. For a scalar-valued estimate, we define the IF at a fixed point. But if the estimate is a function,  we are able to express the change of the function value at every point. Suppose $T(\cdot, F)$ and  $T(\cdot, F^\epsilon)$ are  two function  estimates on the distribution $F_X$ and the contaminated distribution $F^\epsilon$ at $X^\prime$, respectively.  The influence function for  $T(\cdot, F)$ is defined  by
\begin{eqnarray}
IF(\cdot, X ^\prime,  T, F)=\lim_{\epsilon\to 0 }\frac{T(\cdot, F^\epsilon) - T(\cdot, F)}{\epsilon}. \nonumber
\end{eqnarray}
We can estimate the IF using the empirical distribution which is called {\bf empirical IF (EIF)}. Suppose  a sample of size $n$  is  drawn from the empirical distributions  $F_n$. Also let $F^\epsilon_n$  be a contamination model with the empirical data.  The  empirical IF for  $T(\cdot, F_n)$ is defined as
\begin{eqnarray}
IF(\cdot, X ^\prime,  T, F_n)=\lim_{\epsilon\to 0 }\frac{T(\cdot, F^\epsilon_n) - T(\cdot, F_n)}{\epsilon}. \nonumber
\end{eqnarray}

As a  first example, let the  kernel ME, $
T(\cdot, F_X)= \int k_X(\cdot, X)dF_X= E_X[k_X(\cdot, X)]$, where $X\sim F_X.$  The value of the parameter at the contamination model, $F^\epsilon= (1-\epsilon)F_X+\epsilon\Delta_{X^\prime}$ is
\begin{eqnarray}
T(\cdot, F^\epsilon)&=& \int k_X(\cdot, X) \rm{d}[(1-\epsilon)F_X+ \epsilon  \Delta_{X^\prime}] \nonumber\\
&=& (1-\epsilon)\int  k_X(\cdot, X)dF_X +\epsilon  k_X(\cdot, X^\prime)\nonumber\\
&=& (1-\epsilon) T(\cdot,F_X) +\epsilon  k_X(\cdot, X^\prime). \nonumber
\end{eqnarray}
Thus the IF of kernel ME at point $X^\prime$ is given by
\begin{eqnarray}
IF(\cdot, X^\prime, T, F_X) &=&\lim_{\epsilon\to 0 }\frac{T(\cdot, F^\epsilon) - T(\cdot, F_X)}{\epsilon} \nonumber\\
&=&\lim_{\epsilon\to 0 }\biggl[(1-\epsilon)T(\cdot, F_X)+\epsilon k_X(\cdot, X^\prime)-T(\cdot, F_X) \biggr] \nonumber\\
&=&k_X(\cdot, X^\prime)-T(\cdot, F_X)\nonumber\\
&=& k_X(\cdot, X^\prime) - \mb{E}_X[k_X(\cdot, X)], \, \forall\, k_X(\cdot, X^\prime) \in \mc{H}_X. \nonumber
\end{eqnarray}
We can estimate the IF of the  kernel ME with the empirical distribution, $F_n$, at the data points $X_1, X_2, \cdots, X_n \sim F_n$,  at  $X^\prime$ for every point $X$ as
\begin{eqnarray}
 IF(X, X^{\prime}, T, F_n) = k(X, X^\prime) - \frac{1}{n}\sum_{i=1}^n k(X, X_i), \qquad \forall\, k(\cdot, X_i)\in \mc{H}_X,\, \, X\sim F_n, \nonumber
\end{eqnarray}
which is called the EIF of kernel ME.

As a  second example, let the mean of the product of two random variables,  $f(X)$ and $f(Y)$ with   $(X, Y)\in \mc{X}\otimes \mc{Y}$,
  $T(X, Y, F_{XY}) =  \mb{E}_{XY}[f_X(X)f_Y(Y)]$,  for all $f_X\in \mc{H}_X$, and  $f_Y\in \mc{H}_Y$. The value of parameter at the   contamination model at  $(X^\prime, Y^\prime)\in \mc{X}\otimes \mc{Y}$ , $F_{XY}^\epsilon= (1-\epsilon)F_{XY}+\epsilon\Delta_{X^\prime Y^\prime}$ is given by
\begin{eqnarray}
T[F_{XY}^\epsilon]&=&T[(1-\epsilon)F_{XY}+\epsilon\Delta_{X^\prime Y^\prime}]\nonumber \\&=& \int f_X(U) f_Y(V)d[(1-\epsilon)F_{XY}+\epsilon \Delta_{X^\prime Y^\prime}] \nonumber \\
&=&(1-\epsilon)\int f_X(U) f_Y(V)dF_{X Y}+\epsilon\int f_X(U)f_Y(V)d_{\Delta_{X^\prime Y^\prime}}(U, V) \nonumber \\
&=& (1-\epsilon)\int f_X(U) f_Y(V)dF_{XY} +\epsilon f_X(X^\prime)f_Y(Y^\prime)\nonumber\\
&=& (1-\epsilon) T(F_{XY}) +\epsilon f_X(X^\prime)f_Y(Y^\prime) \nonumber.
\end{eqnarray}
Thus the IF of $T(X, Y, F_{XY})$ is given by
\begin{eqnarray}
\label{IFe2}
IF(X, Y, T, F_{XY}) &=&\lim_{\epsilon\to 0 }\frac{T[F_{X Y}^\epsilon] - T(Z, F_{XY})}{\epsilon} \nonumber\\
&=&\lim_{\epsilon\to 0 }\frac{(1-\epsilon)T(Z, F_{XY})+\epsilon f_X(X^\prime) f_Y(Y^\prime)-T(Z, F_{XY})}{\epsilon} \nonumber\\
&=& f_X(X^\prime)f_Y(Y^\prime)-T(X, Y,F_{XY}).
\end{eqnarray}

We can find the IF for a combined statistic given the IF for the statistic itself. The IF of complicated statistics can be calculated with the chain rule,  say $T(F)=\ell(T_1(F),  \cdots, T_s(F))$, that is,
 \begin{eqnarray}
IF_T(X)=\sum_{i=1}^s \frac{\partial \ell}{\partial T_i}IF_{T_i }(X). \nonumber
 \end{eqnarray}
For example, the IF of covariance of two random variables, $f_X(X)$ and $f_Y(Y)$  can be calculated using the above chain rule as
\begin{eqnarray}
T(X, Y, F_{XY}) =\mb{E}_{XY}[f_X(X) f_Y(X)] - \mb{E}_{X}[f_X(X)] \mb{E}_X[f_Y(X)] \nonumber
\end{eqnarray}
for $f_X\in \mc{H}_X$, $f_Y\in \mc{H}_Y$, and $Z=(X,Y)\in \mc{X}\times\mc{Y}$.  

Using  Eq. (\ref{IFe2}) and the reproducing property, the IF of  $T(z, F_{XY})$  with distribution, $F_{XY}$ at  $Z^\prime= (X^\prime, Y^\prime)$  is given by
\begin{eqnarray}
\label{e2iff}
\rm{IF}(\cdot, Z^\prime, T, F_{XY}) &=& f_X(X^\prime)f_Y(Y^\prime)- \mb{E}_{XY}[f_X(X)f_Y(X)]\nonumber \\&-& \mb{E}_{Y}[f_Y(Y)][f_X(X^\prime)-  \mb{E}_{X}[f_X(X)]]- \mb{E}_{X}[f_X(X)][f_Y(Y^\prime)-  \mb{E}_{Y}[f_Y(Y)]] \nonumber \\
&=& [f_X(X^\prime)- \mb{E}_{X}[f_X(X)]][f_Y(Y^\prime)- \mb{E}_{Y}[f_Y(Y)]]- T(Z, F_{XY}) \nonumber \\ 
&=&  \langle k_X(\cdot, X^\prime)-\mc{M}[F_X], f_X \rangle_{\mc{H}_X} \langle k_Y(\cdot,Y^\prime)\mc{M}[F_Y], g \rangle_{\mc{H}_Y} \nonumber \\
&-&\mb{E}_{XY}[ \langle k_X(\cdot, X)-\mc{M}[F_X], f_X \rangle_{\mc{H}_X} \langle k_Y(\cdot,Y)-\mc{M}[F_Y], f_Y \rangle_{\mc{H}_Y}].
 \end{eqnarray}

Letting $f_X= k_X(\cdot, X)$ and $f_Y= k_Y(\cdot, Y),\, \forall\, X\in \mc{X}\, \rm{and}\,  Y\in \mc{Y}$ be two random variables taking values in $\mc{H}_X$ and $\mc{H}_Y$,  the IF of kernel CCO at $Z=(X^\prime, Y^\prime)$ is formulated as
\begin{eqnarray}
\rm{IF}(\cdot, X^\prime, X^\prime, T, F_{XY})= \left[k_X(\cdot, X^\prime)-E_X[k_X(\cdot,X)]\right]\otimes\left[k_Y(\cdot, Y^\prime)-E_Y[k_Y(\cdot,Y)]\right] - \Sigma_{XY},
 \end{eqnarray}
where $k_X(\cdot, X)$,   $k_X(\cdot, X^\prime)$, $k_Y(\cdot, Y)$, and $k_Y(\cdot, Y^\prime)$ are random vectors in $\mc{H}_X$ and $\mc{H}_Y$, respectively.

 Given data points  $(X_1 Y_1), (X_2, Y_2), \cdots, (X_n, Y_n) \in \mc{X}\times \mc{Y}$  from the joint empirical distribution, $F_{nXY}$, for every point $(X_i, Y_i)$, we can estimate the IF of the kernel CCO, called {\bf EIF of kernel CCO} as follows,
\begin{multline}
\widehat{\rm{IF}}(X_i, Y_i, X^\prime, Y^\prime,R, F_{XY})\\ =  [k_X(X_i, X^\prime)-\frac{1}{n}\sum_{b=1}^n k_X(X_i, X_b)] [k_Y(Y_i, Y^\prime)-\frac{1}{n}\sum_{b=1}^n k_Y(Y_i, Y_b)]\\ -[
k_X(X_i, X_d)-\frac{1}{n}\sum_{b=1}^n k_X(X_i, X_b)] [k_Y(Y_i, Y_d)-\frac{1}{n}\sum_{b=1}^n k_Y(Y_i, Y_b)]. \nonumber
\end{multline}

In case of the outliers, the bounded kernels take the values in a range. Thus, the above IFs have the three properties: gross error sensitivity, local shift sensitivity and rejection point only for the bounded kernels. These properties are  not true for the unbounded kernels, for example, linear and polynomial kernels.   The unbounded kernels take the arbitrary values and IFs reflects the bias.  We can make a similar conclusion for the kernel CO.

\subsection{Influence function  of robust kernel (cross-) covariance operator}
To derive the IF of the robust kernel CCO like the robust kernel DE as shown in \citep{Kim-12}, we generalize the definition of robust kernel CCO  to a joint general distribution $\mu_{XY}$,
\begin{eqnarray}
\label{IFR1}
\widehat{\Sigma}_{\mu_{XY}}=\argmin_{\Sigma}\in {\mc{H}_X\otimes\mc{H}_Y} \int \zeta(\| \tilde{\vc{\Phi}}(X) \otimes \tilde{\vc{\Phi}}(Y) - \Sigma\|_ {\mc{H}_X\otimes\mc{H}_Y}) d\mu_{XY}(X,Y).
\end{eqnarray}
 Let $ \widehat{\Sigma}_{RXY}(X, Y; F_{XY}) =\Sigma_{F_{XY}}(X,Y)$, the IF for the robust kernel CCO  at $ (X^\prime, Y^\prime)$ is
\begin{align*}
IF(X,Y, X^\prime, Y^\prime; \widehat{\Sigma}_{RXY}; F_{XY})&= \lim_{\epsilon \to 0}\frac{   \widehat{\Sigma}_{RXY}(X, Y, F^\epsilon_{XY}) -  \widehat{\Sigma}_{RXY}(X, Y, F_{XY})}{\epsilon}\\
 &=\lim_{\epsilon \to 0} \frac{\Sigma_{F_{XY}^\epsilon}-\Sigma_{F_{XY}}}{\epsilon}
\end{align*}
Similarly for the definition of robust kernel CCO,  we generalize the necessary condition $S(\widehat{\Sigma}_{RXY})$  to $S_{F{XY}}(\widehat{\Sigma}_{RXY})$. 
Besides the assumptions $(i) - (iv)$ in Section \ref{sec:rkme}, assume that $\Sigma_{F_{XY}^\epsilon} \to \Sigma_{F_{XY}}$ as $\epsilon \to 0$. We need to find the G\^{a}teaux differentiability of  $S_{F{XY}}$ as in proof of  Lemma {\ref{lemma1}} (in the appendix).  If $\dot{\Sigma}_{F_{XY}} \doteq \lim_{\epsilon \to 0} \frac{\Sigma_{F_{XY}^\epsilon}-\Sigma_{F_{XY}}}{\epsilon}$ exists, the IF of robust kernel CCO is defined as
\[\rm{IF}(X, Y,  X^\prime, Y^\prime, \widehat{\Sigma}_R, F_{XY})= \dot{\Sigma}_{F_{XY}}, \]
where $\dot{\Sigma}_{F_{XY}} \in \mc{H}_X\otimes \mc{H}_Y$ satisfies
\begin{multline}
\label{IFR2}
\Big[\int\varphi(\|\tilde{\vc{\Phi}}(X)\otimes\tilde{\vc{\Phi}}(Y)- \Sigma_{F_{XY}}) \|_{\mc{H}_X\otimes \mc{H}_Y} dF_{XY} \Big] \dot{\Sigma}_{F_{XY}} + \\\int \Big[ \frac{\langle \dot{\Sigma}_{F_{XY}}, \tilde{\vc{\Phi}} (X)\otimes \tilde{\vc{\Phi}}(Y)- \Sigma_{F_{XY}}\rangle_{\mc{H}_X\otimes\mc{H}_Y}}{\|\tilde{\vc{\Phi}} (X)\otimes \tilde{\vc{\Phi}}(Y)-\Sigma_{F_{XY}}\|_ {\mc{H}_X\otimes\mc{H}_Y}^3}\Big. \\ \Big. q(\| \tilde{\vc{\Phi}}(X)\otimes \tilde{\vc{\Phi}}(Y)-\Sigma_{F_{XY}}\|_ {\mc{H}_X\otimes\mc{H}_Y}\|) (\tilde{\vc{\Phi}}(X)\otimes \tilde{\vc{\Phi}} (Y)-\Sigma_{F_{XY}}) \Big] dF_{XY}(X,Y) \\
= (\tilde{\vc{\Phi}} (X^\prime)\otimes \tilde{\vc{\Phi}} (Y^\prime)-\Sigma_{F_{XY}}) (\varphi(\|\tilde{\vc{\Phi}}(X)\otimes \tilde{\vc{\Phi}}(Y)- \Sigma_{F_{XY}}) \|_{\mc{H}_X\otimes \mc{H}_Y}),
\end{multline}
where $q(t)= t \psi^\prime (t)-\psi(t)$. Unfortunately,  Eq. (\ref{IFR2}) has  no closed form solution. By considering the empirical joint distribution, $F_n=F_{nXY}$ instead of the joint distribution, $F_{XY}$, we can find $\dot{\Sigma}_{F_n}$ explicitly. To do this, besides the assumptions $(i) - (iv)$ we assume that $\Sigma_{F_{n}^\epsilon}\to \Sigma_{F_{n}}$  as $\epsilon\to 0$ (satisfied when $J$ is strictly convex) and the extended kernel matrices $\vc{K}_X^\prime$ and $\vc{K}_Y^\prime$ with $(X_i, Y_i)_{i=1}^n \cup (X^\prime, Y^\prime)$ are positive definite. Then, the IF of robust kernel CCO with $(X,Y)$ at $ (X^\prime, Y^\prime)$ is defined as
\[ \rm{IF} (X, Y,  X^\prime, Y^\prime \widehat{\Sigma}_R, F_n)= \sum_{i=1}^n\alpha_i \tilde{k}_X( X, X_i)\tilde{k}_Y(Y, Y_i) + \alpha^\prime \tilde{k}_X( X, X^\prime)\tilde{k}_Y(Y, Y^\prime) \]
where $\alpha^\prime= n\frac{\varphi(\|\tilde{\vc{\Phi}}(X)\otimes\tilde{\vc{\Phi}}(Y)- \Sigma_{F_n}\|_{\mc{H}_X\otimes \mc{H}_Y})}{\gamma}$,  $\gamma= \sum_{i=1}^n\varphi (\|\tilde{\vc{\Phi}} (X_i)\otimes \tilde{\vc{\Phi}} (Y_i)-\Sigma_{F_n}\|_ {\mc{H}_X\otimes\mc{H}_Y})$
and $\alpha=[\alpha_1, \alpha_2, \cdots, \alpha_n]^T$ are the solution of the following system of linear equations:
\begin{multline}
\left\{ \gamma \vc{I}_n+ (\vc{I}_n- \vc{1} \vc{w}^T)^T \vc{Q} (\vc{I}_n- \vc{1} \vc{w}^T)\tilde{\vc{K}}_X\tilde{\vc{K}}_Y\right\}\alpha\\
=- n\varphi(\|\Phi_c(X)\otimes\Phi_c(Y)- \Sigma_{F_n}\|_{\mc{H}_X\otimes \mc{H}_Y})\vc{w}- \alpha^\prime(\vc{I}_n- \vc{1}\vc{w}^T)^T \vc{Q} (\vc{I}_n- \vc{1}\vc{w})\vc{k}_{XY},
\end{multline}
where
$\vc{1}= [1, \cdots, 1]^T$, $\vc{k}_{XY} = [k_X(X^\prime, X_1)k_Y(Y^\prime, Y_1), \cdots, k_X(X^\prime, X_n)k_Y(Y^\prime, Y_n)]$,  $\vc{I}_n$ is a  $n$ ordered  identity matrix,  $\vc{Q}$ is a diagonal matrix with $Q_{ii}= \frac{ q(\|\tilde{\vc{\Phi}} (X_i)\otimes \vc{\Phi} (Y_i)-\Sigma_{F_n}\|_ {\mc{H}_X\otimes\mc{H}_Y})}{ \|\tilde{\vc{\Phi}} (X_i)\otimes \vc{\Phi} (Y_i)-\Sigma_{F_n}\|^3_ {\mc{H}_X\otimes\mc{H}_Y}}$,
and $\vc{w}= [w_1, \cdots, w_n]^T$ gives the weights as in robust kernel CCO.
$\alpha^\prime$ captures the amount of contaminated data in the robust kernel CCO, which is  given as
\[\alpha^\prime = \frac{\varphi(\|\tilde{\vc{\Phi}}(X^\prime)\otimes\tilde{\vc{\Phi}}(Y^\prime)- \Sigma_{F_n}\|_{\mc{H}_X\otimes \mc{H}_Y})}{\frac{1}{n}\sum_{i=1}^n\varphi (\|\tilde{\vc{\Phi}} (X_i)\otimes \tilde{\vc{\Phi}}(Y_i)-\Sigma_{F_n}\|_ {\mc{H}_X\otimes\mc{H}_Y})} \]

For a standard kernel CCO, we have $\varphi \equiv 1$ and $\alpha^\prime=1$, which is in agreement with the IF of standard kernel CCO. The robust loss function  $\zeta$, $\varphi(\|\tilde{\vc{\Phi}}(X^\prime)\otimes\tilde{\vc{\Phi}}(Y^\prime)- \Sigma_{F_n}\|_{\mc{H}_X\otimes \mc{H}_Y})$  can be regarded as a measure of ``inlyingness", with  more inlying points having larger values than $\alpha^\prime <1$. Thus, the robust kernel CCO is less sensitive to outlying points than the standard kernel CCO.

\section{Standard and robust kernel canonical correlation analysis}
In this section, we  review standard kernel CCA and propose the IF  and empirical IF (EIF) of kernel CCA. After that we propose a {\bf robust kernel CCA} method based on robust kernel CO and robust kernel CCO.
\subsection{Standard kernel canonical correlation analysis}
\label{sec:CKCCA}
 Standard kernel CCA has been proposed as a nonlinear extension of linear CCA \citep{Akaho,Lai-00}. Researchers have extended the standard kernel CCA  with  an efficient computational algorithm, i.e., incomplete Cholesky factorization \cite{Back-02}.  Over the last decade,  standard kernel CCA has been used for various tasks \citep{Alzate2008,Hardoon2004,Huang-2009, Ashad-15}. Theoretical results on the convergence of kernel CCA have also been  obtained \citep{Fukumizu-SCKCCA,Hardoon2009}.

The aim of the  standard kernel CCA is to seek  the sets of functions in the RKHS for which the correlation (Corr) of  random variables is maximized. For the simplest case, given two sets of random variables $X$  and $Y$ with  two  functions in the RKHS, $f_{X}(\cdot)\in \mc{H}_X$  and  $f_{Y}(\cdot)\in \mc{H}_Y$, the optimization problem of  the random variables $f_X(X)$ and $f_Y(Y)$ is
\begin{eqnarray}
\label{ckcca1}
\rho =\max_{\substack{f_{X}\in \mc{H}_X,f_{Y}\in \mc{H}_Y \\ f_{X}\ne 0,\,f_{Y}\ne 0}}\mr{Corr}(f_X(X),f_Y(Y)).
\end{eqnarray}
The optimizing functions $f_{X}(\cdot)$ and $f_{Y}(\cdotp)$ are determined up to scale.

Using a  finite sample, we are able to estimate the desired functions. Given an i.i.d sample, $(X_i,Y_i)_{i=1}^n$ from a joint distribution $F_{XY}$, by taking the inner products with elements or ``parameters" in the RKHS, we have features
$f_X(\cdot)=\langle f_X, \vc{\Phi}_X(X)\rangle_{\mc{H}_X}= \sum_{i=1}^na_X^ik_X(\cdot,X_i) $ and
 $f_Y(\cdot)=\langle f_Y, \vc{\Phi}_Y(Y)\rangle_{\mc{H}_Y}=\sum_{i=1}^na_Y^ik_Y(\cdot,Y_i)$, where $k_X(\cdot, X)$ and $k_Y(\cdot, Y)$ are the associated kernel functions for $\mc{H}_X$ and $\mc{H}_Y$, respectively. The kernel Gram matrices are defined as   $\vc{K}_X:=(k_X(X_i,X_j))_{i,j=1}^n $ and $\vc{K}_Y:=(k_Y(Y_i,Y_j))_{i,j=1}^n $.  We need the centered kernel Gram matrices $\vc{G}_X=\vc{C}\vc{K}_X\vc{C}$ and $\vc{G}_Y=\vc{C}\vc{K}_Y\vc{C}$, where $ \vc{C} = \vc{I}_n -\frac{1}{n}\vc{D}_n$ with  $\vc{D}_n = \vc{1}_n\vc{1}^T_n$ and $\vc{1}_n$ is the vector with $n$ ones. The empirical estimate of Eq. (\ref{ckcca1}) is then given by
\begin{eqnarray}
\label{ckcca61}
\hat{\rho}=\max_{\substack{f_{X}\in \mc{H}_X,f_{Y}\in \mc{H}_Y \\ f_{X}\ne 0,\,f_{Y}\ne 0}}\frac{\widehat{\rm{Cov}}(f_X(X),f_Y(Y))}{[\widehat{\rm{Var}}(f_X(X))+\kappa\|f_X\|_{\mc{H}_X}]^{1/2}[\widehat{\rm{Var}}(f_Y(Y))+\kappa\|f_Y\|_{\mc{H}_Y}]^{1/2}}
\end{eqnarray}
where
\begin{align*}
& \widehat{\rm{Cov}}(f_X(X),f_Y(Y))
= \frac{1}{n} \vc{a}_X^T\vc{G}_X\vc{G}_Y \vc{a}_Y= \vc{a}_X^T\vc{G}_X\vc{W}\vc{G}_Y \vc{a}_Y , \\
& \widehat{\rm{Var}}( f_X(X))
=\frac{1}{n} \vc{a}_X^T\vc{G}_X^2 \vc{a}_X= \vc{a}_X^T\vc{G}_X \vc{W} \vc{G}_X \vc{a}_X, \,  \\ &\widehat{\rm{Var}}( f_Y(Y))=\frac{1}{n} \vc{a}_Y^T\vc{G}_Y^2 \vc{a}_Y= \vc{a}_Y^T\vc{G}_Y \vc{W}\vc{G}_Y\vc{a}_Y,
\end{align*}
and  $\vc{W}$ is a diagonal matrix with elements  $\frac{1}{n}$, and  $\vc{a}_{X}$ and $\vc{a}_{Y}$ are the   eigen-direction of  $X$ and $Y$, respectively. The regularized coefficient $\kappa > 0$. Solving the maximization problem in  Eq. (\ref{ckcca61})  is  analogous to solving the  following generalized eigenvalue problem:
\begin{align}
\label{ckcca71}
 \left[\vc{G}_Y\vc{W}\vc{G}_X  (\vc{G}_X \vc{W}\vc{G}_X+\kappa \vc{I})^{-\frac{1}{2}} \vc{G}_X\vc{W}\vc{G}_Y- \rho^2  (\vc{G}_Y \vc{W}\vc{G}_Y+\kappa \vc{I})\right] \vc{a}_{Y}=0 
 \end{align}
\begin{align}
\label{ckcca72}
 \left[ \vc{G}_X\vc{W}\vc{G}_Y  (\vc{G}_Y \vc{W}\vc{G}_Y+\kappa \vc{I})^{-\frac{1}{2}} \vc{G}_Y\vc{W}\vc{G}_X - \rho^2  (\vc{G}_X \vc{W}\vc{G}_X+\kappa \vc{I})\right]\vc{a}_{X}=0 
 \end{align}

It is easy to show  that the eigenvalues of  Eq. (\ref{ckcca71}) and Eq.   (\ref{ckcca72}) are  equal and that the eigenvectors for  any equation   can  be obtained from the other. The square roots of the eigenvalues of Eq.(\ref{ckcca71}) or  Eq. (\ref{ckcca72}) are the estimated kernel CC,  $\hat{\rho}$.   The $\hat{\rho}_j$ is the {\it j}th largest kernel CC and  the  {\it j}th kernel CVs are $\vc{a}_{X}^T\vc{G}_X$, and $\vc{a}_{Y}^T\vc{G}_Y$.

  Standard kernel CCA can be formulated using kernel CCO, which makes the robustness analysis easier.   As in \cite{Fukumizu-SCKCCA}, using the cross-covariance operator of (X,Y), $\Sigma_{XY}: \mc{H}_Y\to \mc{H}_X$  we can reformulate the optimization problem in Eq. (\ref{ckcca1})  as follows:
\begin{eqnarray}
\label{ckcca2}
\sup_{\substack{f_{X}\in \mc{H}_X, f_{Y}\in \mc{H}_Y \\ f_{X}\ne 0,\,f_{Y}\ne 0}}\langle f_X,\Sigma_{XY}f_Y\rangle_{\mc{H}_X}\qquad
\text{subject to}\qquad
\begin{cases}
	\langle f_X, \Sigma_{XX}f_X\rangle_{\mc{H}_X}=1,
\\
\langle f_Y, \Sigma_{YY}f_Y\rangle_{\mc{H}_Y}=1.
\end{cases}
\end{eqnarray}
 As  with linear CCA \citep{Anderson-03}, we can derive the solution of Eq. (\ref{ckcca2}) using the following   generalized eigenvalue problem.
\begin{eqnarray}
\label{ckcca3}
\begin{cases}
\Sigma_{XY}f_X - \rho\Sigma_{YY} f_Y = 0,
\\
\Sigma_{XY}f_Y - \rho\Sigma_{XX}f_X = 0.
\end{cases} \nonumber
\end{eqnarray}
 The  eigenfunctions of Eq. (\ref{ckcca3}) correspond to the largest eigenvalue,   which is  the solution to the kernel CCA problem.  After some simple calculations, we  reset the solution as
\begin{eqnarray}
\label{ckcca4}
\begin{cases}
(\Sigma_{XY} \Sigma_{YY}^{-1}\Sigma_{XY} - \rho^2 \Sigma_{XX})f_X = 0,\\
(\Sigma_{XY} \Sigma_{XX}^{-1}\Sigma_{XY}-\rho^2 \Sigma_{YY})f_Y = 0.
\end{cases}
\end{eqnarray}

It is known that the inverse of an operator may not exist. Even if it exists, it may not be continuous in general \citep{Fukumizu-SCKCCA}. We can derive kernel CC using the correlation operator
$\Sigma_{YY}^{- \frac{1}{2}}\Sigma_{YX} \Sigma_{XX}^{- \frac{1}{2}}$, even when $\Sigma_{XX}^{- \frac{1}{2}}$ and  $\Sigma_{YY}^{- \frac{1}{2}}$ are not proper operators. The potential danger  is that it might overfit, which is why introducing $\kappa$ as a regularization coefficient would be helpful.   Using the regularized coefficient $\kappa > 0$, the empirical estimators of Eq. (\ref{ckcca2}) and Eq. (\ref{ckcca4}) are
\begin{eqnarray}
\sup_{\substack{f_{X}\in \mc{H}_X,f_{Y}\in \mc{H}_X \\ f_{X}\ne 0,\,f_{Y}\ne 0}}\langle f_Y,\hat{\Sigma}_{YX}f_X\rangle_{\mc{H}_Y}\qquad
\text{subject to}\qquad
\begin{cases}
\langle f_X, (\hat{\Sigma}_{XX} + \kappa\vc{I}) f_X\rangle_{\mc{H}_X} = 1,
\\
\langle f_Y, (\hat{\Sigma}_{YY}+\kappa\vc{I})f_Y\rangle_{\mc{H}_Y} = 1,
\end{cases}
\label{ckcca73}
\end{eqnarray}
and 
\begin{eqnarray}
\label{ckcca8}
\begin{cases}
(\hat{\Sigma}_{XY} (\hat{\Sigma}_{YY} + \kappa\vc{I})^{-1}\hat{\Sigma}_{XY} - \rho^2 (\hat{\Sigma}_{XX}+\kappa\vc{I}))f_X = 0,
\\
(\hat{\Sigma}_{YX} (\hat{\Sigma}_{XX} + \kappa\vc{I})^{-1}\hat{\Sigma}_{YX} - \rho^2 (\hat{\Sigma}_{YY}+\kappa\vc{I}))f_Y = 0,
\end{cases}
\end{eqnarray}
respectively.

   Now we calculate a finite rank operator  $ \mb{B}_{YX} = (\hat{\Sigma}_{{YY}} + \kappa \vc{I})^{- \frac{1}{2}} \hat{\Sigma}_{{YX}} (\hat{\Sigma}_{{XX}} + \kappa\vc{I})^{- \frac{1}{2}}$.  For $\kappa > 0$,  the square roots of the {\it j}-th   eigenvalue  of $\mb{B}_{YX}$ are the   {\it j}-th kernel CC, $\rho_j$.   The unit  eigenfunctions of   $\mb{B}_{YX}$  corresponding to the  {\it j}th eigenvalues  are $\hat{\nu}_{jX}\in \mc{H}_X$ and $\hat{\nu}_{j_Y}\in \mc{H}_Y$. The {\it j}th ($j= 1, 2, \cdots, n$)   kernel CVs are  
\[ \hat{f}_{jX}(X) = \langle \hat{f}_{jX}, \tilde{k}_X(\cdot, X)\rangle \,\rm{and}\,\hat{f}_{jY}(X) = \langle \hat{f}_{jY}, \tilde{k}_Y(\cdot, Y) \rangle\]
 where   $\hat{f}_{jX} =  (\hat{\Sigma}_{{XX}} + \kappa\vc{I})^{-\frac{1}{2}}\hat{\nu}_{jX}$ and   $\hat{f}_{jY} =  (\hat{\Sigma}_{{YY}} + \kappa\vc{I})^{-\frac{1}{2}}\hat{\nu}_{jY}.$

The generalized eigenvalue problem  in  Eq. (\ref{ckcca4})  can be formulated as a simple eigenvalue problem. Using the  {\it j}-th eigenfunction in  the first equation  of Eq. (\ref{ckcca4})  we have 
\begin{eqnarray}
\label{ckcca5}
(\Sigma_{XX}^{- \frac{1}{2}} \Sigma_{XY} \Sigma_{YY}^{-1}\Sigma_{YX}\Sigma_{XX}^{- \frac{1}{2}} - \rho_j^2I) \Sigma_{XX}^{\frac{1}{2}}f_{jX}&=&0 \nonumber \\
\Rightarrow (\Sigma_{XX}^{-\frac{1}{2}} \Sigma_{XY} \Sigma_{YY}^{-1}\Sigma_{YX}\Sigma_{XX}^{- \frac{1}{2}} - \rho_j^2I)e_{jX}& = &0
\end{eqnarray}
where $e_{jX} = \Sigma_{XX}^{\frac{1}{2}}f_{jX}$.
\subsection{Influence function of the standard kernel canonical correlation analysis}
\label{sec:IFKCCA}
By using  the   IF   of  kernel PCA, linear PCA and   linear CCA,   we  can derive  the IF of kernel CCA (kernel CC and kernel CVs).
For simplicity, let us define $ \tilde{f}_X(X)=\langle f_X,  \tilde{k}_X (\cdot, X)$,  $\mb{L}_{jX}= \Sigma_{XX}^{- \frac{1}{2}}(\Sigma_{XX}^{- \frac{1}{2}} \Sigma_{XY} \Sigma_{YY}^{-1} \Sigma_{YX}\Sigma_{XX}^{- \frac{1}{2}}-\rho^2_j\vc{I})^{-1}\Sigma_{XX}^{- \frac{1}{2}}$, and  $\mb{L}_{jY}= \Sigma_{YY}^{- \frac{1}{2}}(\Sigma_{YY}^{- \frac{1}{2}} \Sigma_{YX} \Sigma_{XX}^{-1} \Sigma_{XY}\Sigma_{YY}^{- \frac{1}{2}}-\rho^2_j\vc{I})^{-1}\Sigma_{YY}^{- \frac{1}{2}}$.

\begin{thm}
\label{TIFKCCA}
 Given two sets of random variables $(X, Y)$ having the  distribution  $F_{XY}$ and the  { \it j}-th kernel CC ( $\rho_j$) and kernel  CVs ($f_{jX}(X)$ and $f_{jX}(Y)$), the  influence functions of  kernel CC  and kernel CVs  at $Z^\prime = (X^\prime, Y^\prime)$ are
\begin{multline}
\rm{IF} (Z^\prime, \rho_j^2)= - \rho_j^2 \tilde{f}_{jX}^2(X^\prime) + 2 \rho_j \tilde{f}_{jX}(X^\prime) \tilde{f}_{jY}(Y^\prime)  - \rho_j^2 \tilde{f}_{jY}^2(Y^\prime), \nonumber\\
\rm{IF} (\cdot, Z^\prime, f_{jX}) = -\rho_j [\tilde{f}_{jY}(Y^\prime) - \rho_j \tilde{f}_{jX}(X^\prime)]\mb{L}_{jX} \tilde{k}_X (\cdot, X^\prime) -  [\tilde{f}_{jX}(X^\prime)  - \rho_j \tilde{f}_{jY}(Y^\prime)]\mb{L}_{jX}\Sigma_{XY}\Sigma^{-1}_{YY} \tilde{k}_Y(\cdot,  Y^\prime)\nonumber\\\qquad\qquad\qquad\qquad\qquad\qquad\qquad\qquad\qquad\qquad\qquad\qquad+\frac{1}{2}[1- \tilde{f}^2_{jX}(X^\prime)]f_{jX}, \nonumber\\
 \rm{IF} (\cdot, Z^\prime, f_{jY})
= -\rho_j [\tilde{f}_{jX}(X^\prime) - \rho_j \tilde{f}_{jY}(Y^\prime)]\mb{L}_{jY} \tilde{k}_Y(\cdot, Y^\prime)-  [\tilde{f}_{jY}(Y^\prime)- \rho_j \tilde{f}_{jX}(X^\prime)]\mb{L}_{jY}\Sigma_{YX}\Sigma^{-1}_{XX} \tilde{k}_Y(\cdot, Y^\prime) \nonumber\\ +\frac{1}{2}[1- \tilde{f}^2_{jY}(Y^\prime)]f_{jY}. \nonumber
\end{multline}
\end{thm}
The above theorem  has been  proved on the basis of previously established ones, such as the IF of  linear  PCA \citep{Tanaka-88, Tanaka-89}, the IF of linear CCA \citep{Romanazii-92},  and the IF of  kernel PCA, respectively. To do this,  we  convert the generalized eigenvalue problem of kernel CCA  into a simple eigenvalue problem. First, we need to find the IF of  $ \Sigma_{XX}^{- \frac{1}{2}} \Sigma_{XY} \Sigma_{YY}^{-1}\Sigma_{YX}\Sigma_{XX}^{- \frac{1}{2}}$, henceforth the IF of $\Sigma_{YY}^{-1},  \Sigma_{XX}^{\frac{1}{2}}$ and $\Sigma_{XY}$.

 Using the above  result,  we can establish some properties of kernel CCA: robustness, asymptotic consistency and its standard error. In addition, we  are able to identify the outliers based on the influence of the data. All notations and proof are explained in the appendix.

  The IF of inverse covariance operator exists  only for the finite dimensional RKHS. For infinite dimensional RKHS,  we can find the IF of $\Sigma_{XX}^{- \frac{1}{2}}$  by introducing  a regularization term as follows
\begin{multline}
\rm{IF}(\cdot, X^\prime,
(\Sigma_{XX} + \kappa\vc{I})^{- \frac{1}{2}}) =\nonumber\\ \frac{1}{2} [(\Sigma_{XX}+\kappa\vc{I})^{- \frac{1}{2}}- (\Sigma_{XX}+\kappa\vc{I})^{- \frac{1}{2}}\tilde{k}_X(\cdot, X^\prime)\otimes  \tilde{k}_X(\cdot, X^\prime)(\Sigma_{XX} + \kappa\vc{I})^{- \frac{1}{2}}], \nonumber
\end{multline}
which gives the empirical estimator.

Let $(X_i, Y_i)_{i=1}^n$ be a sample from the  empirical joint distribution $F_{nXY}$. The EIF (IF based on empirical distribution) of kernel CC and  kernel CVs  at $(X^\prime, Y^\prime)$ for all points $ (X_i, Y_i)$ are $\rm{EIF} (X_i, Y_i, X^\prime, Y^\prime, \rho_j^2) = \widehat{\rm{IF}} ( X^\prime, Y^\prime,  \hat{\rho}_j^2)$, $\rm{EIF} (X_i, Y _i,  X^\prime, Y^\prime, f_{jX}) = \widehat{\rm{IF}} (\cdot, X^\prime, Y^\prime, f_{jX})$, and $\rm{EIF} (X_i, Y_i,   X^\prime, Y^\prime, f_{jY}) = \widehat{\rm{IF}} (\cdot,  X^\prime, Y^\prime, \widehat{f}_{jY})$, respectively.

For the bounded kernels, the IFs defined in Theorem \ref{TIFKCCA} have  three properties: gross error sensitivity, local shift sensitivity, and rejection point. But for unbounded kernels, say a linear, polynomial and so on,  the IFs are not bounded. Consequently, the results of standard kernel CCA using the bounded kernels are less sensitive than  the  standard kernel CCA using the unbounded kernels.  In practice, standard kernel CCA is sensitive to the contaminated data even with the  bounded kernels \citep{Ashad-10}.

\subsection{Robust kernel canonical correlation analysis}
\label{sec:RKCCA}
In this section, we propose  a {\bf robust kernel CCA} method based on the  robust kernel CO and the  robust kernel CCO. While many robust linear CCA  methods have been proposed to show  that linear CCA methods cannot fit the bulk of the data  and have  points deviating from the original pattern  for further investment \citep{Adrover-15, Ashad-10}, there are no  well-founded robust methods of  kernel CCA. The  standard kernel CCA  considers the same weights for each data point, $\frac{1}{n}$,  to estimate kernel CO and kernel CCO, which is the solution of an empirical risk optimization problem when using the quadratic loss function. It is known that the least square loss function is not a robust loss function. Instead, we can solve  an empirical risk optimization problem using the robust least square loss function where  the weights are  determined based on  KIRWLS. We need robust centered kernel Gram matrices of $X$ and $Y$ data.  The  centered  robust kernel Gram matrix of $X$ is,   $\vc{G}_{RX}=\vc{C_{R}}\vc{K}_X\vc{C_{R}}$  where $ \vc{C}_R = \vc{I}_n - \vc{1}_n\vc{w}^T_n$, $\vc{1}_n$ is the vector with $n$ ones and $\vc{w}$ is a weight vector of robust kernel ME, $\mc{M}_X$. Similarly,  we can calculate  $\vc{G}_{RY}$ for $Y$.  After getting robust kernel CO and kernel CCO, they are used in standard kernel CCA, which we call {\bf robust kernel CCA}.  The empirical estimate of Eq. (\ref{ckcca1}) is then given by
\begin{eqnarray}
\label{ckcca6}
\hat{\rho}_{rkcc}=\max_{\substack{g_{X}\in \mc{H}_X,g_{Y}\in \mc{H}_Y \\ g_{X}\ne 0,\,g_{Y}\ne 0}}\frac{\widehat{\rm{Cov}_R}(g_X(X),g_Y(Y))}{[\widehat{\rm{Var}_R}(g_X(X))+\kappa\|g_X\|_{\mc{H}_X}]^{1/2}[\widehat{\rm{Var}_R}(g_Y(Y))+\kappa\|g_Y\|_{\mc{H}_Y}]^{1/2}} \nonumber
\end{eqnarray}
 with for all $g_X\in {\mc{H}}_X$,  $g_Y\in {\mc{H}}_Y$ and
\begin{align*}
& \widehat{\rm{Cov}_R}(g_X(X),g_Y(Y))
=  \vc{b}_X^T\vc{G}_{RX}\vc{W}_{XY}\vc{G}_{RY} \vc{b}_Y , \\
& \widehat{\rm{Var}_R}( g_X(X))
= \vc{b}_X^T\vc{G}_{RX} \vc{W}_{XX} \vc{G}_{RX} \vc{b}_X, \,  \\ &\widehat{\rm{Var}_R}( g_Y(Y))= \vc{b}_Y^T\vc{G}_{RY} \vc{W}_{YY} \vc{G}_{RY}\vc{b}_Y,
\end{align*}
 where $\vc{W}_{XY}$,  $\vc{W}_{XX}$, and $\vc{W}_{YY}$ are   diagonal matrices with elements corresponding to the  weights of  robust kernel CCO, and kernel COs, respectively.  Also  $\vc{b}_{X}$ and $\vc{b}_{Y}$ are the   eigen-direction  of   $X$ and $Y$, respectively. As in  Eq. (\ref{ckcca71}), we can   solve the  maximization problem of Eq. (\ref{ckcca6}) as an  eigenvalue problem. 
Let  $\Sigma_{RXY}$, $\Sigma_{RXX}$, and $\Sigma_{RYY}$ be the  robust kernel CCO, robust kernel CO of $X$, and   robust kernel CO of $Y$, respectively. Like standard kernel CCA, the  robust empirical estimators of Eq. (\ref{ckcca2}) and Eq. (\ref{ckcca4}) are
\begin{eqnarray}
\sup_{\substack{f_{X}\in \mc{H}_X,f_{Y}\in \mc{H}_X \\ f_{X}\ne 0,\,f_{Y}\ne 0}}\langle f_Y,\hat{\Sigma}_{RXY}f_X\rangle_{\mc{H}_Y}\qquad
\text{subject to}\qquad
\begin{cases}
\langle f_X, (\hat{\Sigma}_{RXX} + \kappa\vc{I}) f_X\rangle_{\mc{H}_X} = 1,
\\
\langle f_Y, (\hat{\Sigma}_{RYY}+\kappa\vc{I})f_Y\rangle_{\mc{H}_Y} = 1,
\end{cases}
\label{rkcca73}
\end{eqnarray}
and 
\begin{eqnarray}
\label{rkcca8}
\begin{cases}
(\hat{\Sigma}_{RXY} (\hat{\Sigma}_{RYY} + \kappa\vc{I})^{-1}\hat{\Sigma}_{RXY} - \rho^2 (\hat{\Sigma}_{RXX}+\kappa\vc{I}))f_X = 0,
\\
(\hat{\Sigma}_{RXY} (\hat{\Sigma}_{RXX} + \kappa\vc{I})^{-1}\hat{\Sigma}_{RXY} - \rho^2 (\hat{\Sigma}_{RYY}+\kappa\vc{I}))f_Y = 0,
\end{cases}
\end{eqnarray}
respectively.
Figure \ref{RobustKCCA} presents  a detailed algorithm  of the proposed methods (all steps are similar to standard kernel CCA except the  first one). This method  is designed for  contaminated data, and  the principles we describe also apply to the  kernel methods, which must deal with the issue of kernel CO and kernel CCO.

  It is well-known that robust methods have higher time complexity than the standard methods. At each update of the robust kernel CO or robust kernel CCO, we need to store the $n\times n$ matrix. The memory complexity of robust kernel CCA is then $\mc{O}(n^3)$. A naive implementation of the algorithm  in Figure  \ref{RobustKCCA}  would therefore require  $\mc{O}(n^3h))$ operations (the time complexity), where $h$ is the number of iterations. The spectrum of Gram matrices tends to show rapid decay, and   low-rank approximations of Gram matrices can  often provide sufficient fidelity for the needs of  kernel-based algorithms \citep{Drineas-05, Schlkof-book, Back-02}. By assuming that the outliers have  a similar effect on marginal distribution and the joint distribution, we can also reduce the memory complexity and time complexity. Under this assumption, we estimate the weight of kernel CCO and consider this weight for kernel CO of $X$ and $Y$ data.

\begin{figure}
\begin{em}
\noindent\fbox{%
    \parbox{\textwidth}{%
Input: $D=\{(X_1,Y_1), (X_2,Y_2), \ldots, (X_n,Y_n) \}$ in $\mb{R}^{m_1\times m_2}$.
\begin{enumerate}
\item Calculate the robust kernel cross-covariance operator, $\hat{\Sigma}_{RXY}$ and kernel covariance operators,  $\hat{\Sigma}_{RXX}$  and   $\hat{\Sigma}_{RYY}$ using algorithm in  Figure \ref{Robust.cross.covariance.M}.
\item Find $ \mb{B}_{YX} = (\hat{\Sigma}_{RYY} + \kappa \vc{I})^{- \frac{1}{2}} \hat{\Sigma}_{RYX} (\hat{\Sigma}_{RXX} + \kappa\vc{I})^{- \frac{1}{2}}$
\item For $\kappa > 0$, we have $\rho^2_j$  the largest eigenvalue of $\mb{B}_{YX}$ for $j=1, 2,\cdots, n$.
\item  The unit  eigenfunctions of   $\mb{B}_{YX}$  corresponding to the  $j$th eigenvalues  are $\hat{\xi}_{jX}\in \mc{H}_X$ and $\hat{\xi}{j_Y}\in \mc{H}_Y$
\item The \it {j}th ($j= 1, 2, \cdots, n$)  robust kernel canonical variates are given by
\[ \hat{g}_{jX}(X) = \langle \hat{g}_{jX}, \tilde{k}_X(\cdot, X)\rangle \,\rm{and}\,\hat{g}_{jY}(X) = \langle \hat{g}_{jY}, \tilde{k}_Y(\cdot, Y) \rangle\]
 where   $\hat{g}_{jX} =  (\hat{\Sigma}_{RXX} + \kappa\vc{I})^{-\frac{1}{2}}\hat{\xi}_{jX}$ and   $\hat{g}_{jY} =  (\hat{\Sigma}_{RYY} + \kappa\vc{I})^{-\frac{1}{2}}\hat{\xi}_{jY}$
\end{enumerate}
Output:  the robust kernel CCA
}
}
\caption{The algorithm for estimating robust  kernel CCA.}
\label{RobustKCCA}
\end{em}
\end{figure}

\section{Experiments}
\label{sec:Exp}
We conducted experiments on both the synthetic and real data sets. We generated two types of simulated  data: ideal data  and those with $5\%$ of  contamination.  The description of real data sets are in Sections \ref{Sec:apr}. The five synthetic data sets are as follows:

\textbf {Three circles of structural data (TCSD):}
Data are generated along three circles of different radii with small noise:
\begin{equation}
X_i = r_i \begin{pmatrix} \cos(\vc{Z}_i) \\ \sin(\vc{Z}_i) \end{pmatrix}
+ \epsilon_i, \nonumber
\end{equation}
where $r_i=1$, $0.5$ and $0.25$, for $i=1,\ldots,n_1$, $i=n_1+1,\ldots,n_2$, and $i=n_2+1,\ldots, n_3$, respectively, $\vc{Z}_{i}\sim U[-\pi, \pi]$ and $\epsilon_i\sim \mc{N} (0, 0.01\,I_2)$ independently for the ideal data and  $\vc{Z}_{i}\sim U[-10, 10]$ for the contaminated data.

\textbf {Sine function of structural data (SFSD):}
 1500 data points are generated along the sine function with small noise:
\begin{equation}
X_i = \begin{pmatrix} \vc{Z}_i \\ 2\sin(2\vc{Z}_i)\\
\vdots \\10\sin(10\vc{Z}_i) \end{pmatrix}
+ \epsilon_i,  \nonumber
\end{equation}
where $\vc{Z}_{i}\sim U[-2\pi,0]$ and $\epsilon_i\sim \mc{N} (0, 0.01\,I_{10})$ independently for the  ideal data and $\epsilon_i\sim \mc{N} (0, 10\,I_{10})$ for the contaminated data.

\textbf {Multivariate Gaussian structural data (MGSD): }
Given  multivariate normal data, $\vc{Z}_i\in\mb{R}^{12} \sim \vc{N}(\vc{0},\Sigma)$ ($i= 1, 2, \ldots, n$), where  $\Sigma$ is the same as in \citep{Ashad-15}. We  divide $\vc{Z}_i$ into two sets of variables ($\vc{Z}_{i1}$,$\vc{Z}_{i2}$), and use the first six variables of $\vc{Z}_i$ as $X$ and perform  the $\log$  transformation of the absolute value of the remaining  variables ($\log_e(|\vc{Z}_{i2}|))$) as $Y$. For the contaminated data  $\vc{Z}_i\in\mb{R}^{12} \sim \vc{N}(\vc{1},\Sigma)$ ($i= 1,2,\ldots, n$).

\textbf {Sine and cosine function structural data (SCFSD):}
 We use uniform   marginal distribution, and transform the data by two periodic $\sin$ and $\cos$ functions to make two sets $X$ and $Y$, respectively, with additive Gaussian noise:
$Z_i\sim U[-\pi,\pi], \,\eta_i\sim N(0,10^{-2}),~\,i=1,2,\ldots, n,
 X_{ij}=\sin(jZ_i)+\eta_i,\,  Y_{ij} = \cos(jZ_i)+\eta_i, j=1,2,\ldots, 100.$
For the contaminated model $\eta_i\sim N(1,10^{-2})$.

\textbf {SNP and fMRI structural data (SMSD):}
Two data sets of SNP data X with $1000$ SNPs and fMRI data Y with 1000 voxels were simulated. To correlate the SNPs with the voxels, a  latent model is used   as in \citep{Parkhomenko-09}). For simulation of  contamination,  we consider the signal level, $0.5$  and  noise level, $1$  to $10$ and $20$, respectively.

In our experiments, first, we compare standard and robust  kernel covariance operators. After that, the robust kernel CCA is compared with the standard kernel CCA. For the Gaussian kernel we use the median of the  pairwise distance as a  bandwidth  and  for the Laplacian kernel we set the bandwidth equal to 1. As shown in \citep{Hardoon2004},  we can  optimize  the regularization parameter but our goal is the robustness issue of  different methods. Thus the regularization parameter in standard  kernel CCA and robust kernel CCA is  fixed as  $\kappa= 10^{-5}$.  In robust methods, we consider  Huber's loss function with the constant, $c$, equal to the median of error.

\subsection{Results  of kernel CCO and robust CCO}
We evaluate the performance of kernel CO  and robust kernel CO in two different settings. First, we check the accuracy of both operators by considering the  kernel  CO (KCO)  with large data (say a population kernel CO of size $N$) and kernel CO with small data (say a sample kernel CO of size $n$). Now, we can estimate the distance between sample kernel CO  and population kernel CO as $\|\widehat{\Sigma}_{XX} -\Sigma_{XX}\|^2_{\mc{H}_X\otimes\mc{H}_X}= \|\widehat{\Sigma}_{XX}\|^2_{\mc{H}_X\otimes\mc{H}_X}- 2\langle \widehat{\Sigma}_{XX},  \Sigma_{XX}\rangle_{\mc{H}_X\otimes\mc{H}_X} + \|\Sigma_{XX}\|^2_{\mc{H}_X\otimes\mc{H}_X} $ and similarly for the robust kernel CO (RKCO). Thus,  the performance  measures of the kernel CO and   robust  kernel CO  estimators are defined as
\begin{multline}
\eta_{KCO}=\frac{1}{n^2}\sum_{i=1}^n\sum_{j=1}^nk_X(X_i,X_j)^2 - 2\frac{1}{Nn}\sum_{i=1}^n\sum_{J=1}^Nk_X(X_i,X_J)^2+ \frac{1}{N^2}\sum_{I=1}^N\sum_{J=1}^Nk_X(X_I,X_J)^2,\nonumber \end{multline}
and
\begin{multline}
\eta_{RKCO}=\sum_{i=1}^n\sum_{j=1}^n w_iw_jk_X(X_i,X_j)^2- 2\frac{1}{N}\sum_{i=1}^n w_i\sum_{J=1}^Nk_X(X_i,X_J)^2 + \frac{1}{N^2}\sum_{I=1}^N\sum_{J=1}^Nk_X(X_I,X_J)^2, \nonumber
\end{multline}
 respectively.

\begin{figure}
\begin{center}
\includegraphics[width=\columnwidth, height=12cm]{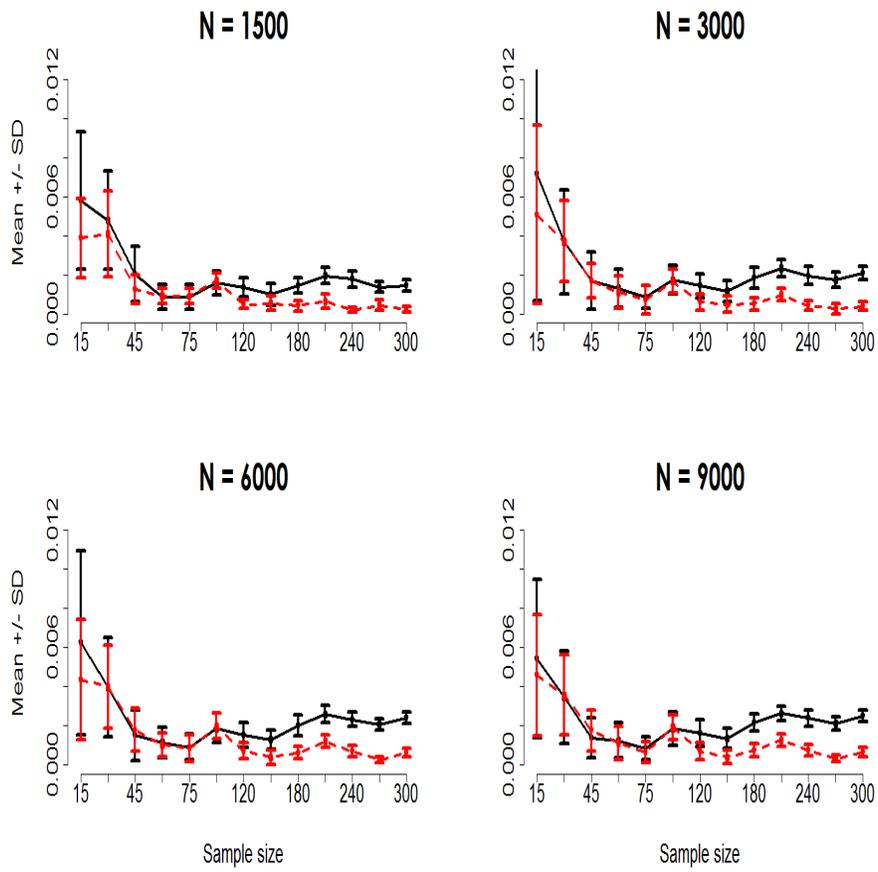}
\caption{Accuracy measure of the  standard kernel covariance operator (solid line) and the robust kernel covariance operator (dash-dotted line).}
\label{Fcons}
\end{center}
\end{figure}

In theory, the above two equations become zero for large population size, $N$, with the sample size, $n\to N$. To do this, we consider the synthetic data, TCSD with  $N\in\{1500, 3000, 6000, 9000\}$ and   $n\in\{15, 30, 45, 60, 90, 120, 150, 180, 210, 240, 270, 300\}$ ($n = n_1 + n_2 + n_3$). For each  $n$, we take  $5\%$ CD. We  repeated the process for  $100$ samples  to confirm our findings.    The results (mean with standard error) were plotted in Figure \ref{Fcons}.  These figures show that both estimators give similar performances for small sample sizes, but for large sample sizes  the   robust estimator (i.e., robust kernel CO) shows much better results than the kernel CO estimate at all  population sizes.

In addition, we compare kernel CO and robust  kernel CO  estimators using five kernels: linear (Poly-1), a polynomial with degree $2$ (Poly-2) and polynomial with degree $3$ (Poly-3), Gaussian and Laplacian on two synthetic data sets: TCSD and SFSD. To measure the performance, we use two matrix norms: Frobenius norm (F) and maximum modulus of all the elements (M) \citep{Sequeira-11}.   We calculate the ratio  between  ideal model   and  contaminated model for the kernel CO. The ratio becomes  zero if the estimator is not sensitive to contaminated data.   For both estimators, kernel CO and robust kernel CO,  we use the following  performance measures,
\[\eta_{KCOR} = \left| 1- \frac{\| \widehat{\Sigma_{XX}}^{ID}\|}{\|\widehat{\Sigma_{XX}}^{CD} \|} \right|,\]
and 
 \[\eta_{RKCOR} = \left| 1- \frac{\| \widehat{\Sigma_{RXX}}^{ID}\|}{\|\widehat{\Sigma_{RXX}}^{CD} \|} \right|,\]
respectively.  We repeated the experiment  for $100$ samples with sample size, $n=1500$. The results (mean $\pm$ standard deviation) for kernel CO (standard) and robust kernel CO (Robust)  are tabulated in  Table \ref{NRCRcov}.  From this table, it is clear that  the  robust estimator  performs better than the standard estimator in all cases. Moreover, both estimators  using  Gaussian and Laplacian kernels are less sensitive than all   polynomial kernels.
\begin{table}
\begin{center}
\caption {Mean  and  standard deviation of the  measure of  kernel covariance operator (Standard) and  robust kernel covariance operator (Robust).}
\label{tbl:norm}
 \begin{tabular}{llcccc} \hline
&Data&\multicolumn{2}{c}{TCSD}&\multicolumn{2}{c}{SFSD}\\ \cline{2-6}
Measure&Kernel&Standard&Robust &Standard&Robust    \\ \hline
&Poly-1&$0.9874\pm 0.0017$&$0.8963\pm 0.0069$&$0.3067\pm  0.1026$&$  0.1669 \pm 0.0626$\tabularnewline
&Poly-2&$1.0000\pm 0.0000$&$0.9863\pm 0.0020$&$0.9559\pm 0.0622$&$  0.5917  \pm 0.1598$\tabularnewline
$\| \widehat{\Sigma}_{XX} \|_F$&Poly-3&$1.0000 \pm 0.0000$&$0.9996\pm 0.0001$&$0.9973\pm 0.0094 $&$ 0.8793  \pm 0.1067$\tabularnewline
&Gaussian&$0.1153\pm 0.0034$&$0.1181\pm 0.0039$&$0.1174\pm 0.0266 $&$ 0.1059  \pm 0.0258$\tabularnewline
&Laplacian&$0.1420\pm 0.0032$&$0.1392\pm 0.0035$&$0.1351\pm 0.0459 $&$ 0.1280  \pm 0.0366$\tabularnewline \hline
&Poly-1&$0.9993\pm 0.0001$&$0.9940\pm 0.0005$& $0.8074 \pm 0.0838$&$  0.6944\pm  0.1118$\tabularnewline
&Poly-2&$1.0000\pm 0.0000$&$0.9996\pm 0.0001$&$0.9921 \pm  0.0122 $&$ 0.9070 \pm 0.0703$\tabularnewline
$\| \widehat{\Sigma}_{XX} \|_M$&Poly-3&$1.0000\pm 0.0000$&$1.0000\pm 0.0000$&$0.9994 \pm 0.0020 $&$ 0.9709 \pm 0.0344$\tabularnewline
&Gaussian&$0.1300\pm 0.0133$&$0.1028\pm 0.0038$&$0.1065 \pm 0.0583 $&$  0.0735 \pm 0.0370$\tabularnewline
&Laplacian&$0.1877\pm 0.0053$&$0.1474\pm 0.0042$&$0.1065 \pm 0.0583 $&$  0.0735 \pm 0.0370$\tabularnewline
\hline
\end{tabular}
\label{NRCRcov}
\end{center}
\end{table}

\subsection{Visualizing influential subject using standard kernel CCA and robust kernel CCA}
\label{Sec:Visu}
We evaluated the performance of  the proposed method for three different settings. First, we compared  robust kernel CCA with the standard kernel CCA using Gaussian kernel (same bandwidth and regularization). To measure the influence, we calculated the  ratio of IF for kernel CC between  ideal data  and  contaminated data. We also calculated a  similar measure  for the kernel CV.  Based on these ratios, we defined two performance measures on kernel  CC and kernel CVs
\begin{eqnarray}
\eta_{\rho} &=&\left | 1- \frac{\|EIF(\cdot, \rho^2) ^{ID}\|_F}{\|EIF(\cdot, \rho^2)^{CD}\|_F}\right| \qquad \rm{and} \\
\eta_{f} &=& \left| 1- \frac{\|EIF(\cdot, f_X)^{ID}- EIF(\cdot,f_Y)^{ID}\|_F}{\|EIF(\cdot, f_X) ^{CD}-EIF(\cdot, f_Y)^{CD}\|_F}\right|,\nonumber
\end{eqnarray}
respectively. For any  method, that does not depend on the contaminated data, the above measures, $\eta_{\rho}$  and $\eta_{f}$, should  be approximately zero. In other words, the best methods should give small values. To compare, we considered three simulated data sets: MGSD, SCFSD, SMSD with three sample sizes, $n\in \{ 100, 500, 1000\}$. For each sample size, we repeated the experiment for $100$ samples.  Table \ref{tbl:ifnorm} presents the results (mean $\pm$ standard deviation) of the standard kernel CCA and robust kernel CCA. From this table, we observed that the robust kernel CCA outperforms the standard kernel CCA in all cases.

\begin{table}
 \begin{center}
\caption {Mean and standard deviation of the measures, $\eta_{\rho}$ and $\eta_{f}$ of the standard kernel CCA (Standard) and  robust kernel CCA (Robust).}
\label{tbl:ifnorm}
 \begin{tabular}{llcccccccc} \hline
&Measure &\multicolumn{2}{c}{$\eta_{\rho}$}&\multicolumn{2}{c}{$\eta_{f}$}\\ \cline{3-6}
Data&n&$Standard$&$Robust$&$Standard$&$Robust$  \\ \hline
&$100$&$1.9114\pm 3.5945$&$ 1.2445\pm 3.1262$&$ 1.3379\pm 3.5092$&$ 1.3043\pm 2.1842$\tabularnewline
MGSD&$500$&$ 1.1365\pm 1.9545$&$ 1.0864\pm 1.5963$&$ 0.8631 \pm 1.3324 $&$0.7096\pm 0.7463$\tabularnewline
&$1000$&$ 1.1695\pm 1.6264$&$ 1.0831\pm 1.8842$&$ 0.6193 \pm 0.7838$&$ 0.5886\pm 0.6212$\\ \hline
&$100$&$0.4945\pm 0.5750$& $0.3963\pm 0.4642$& $1.6855\pm 2.1862$& $0.9953\pm 1.3497$\tabularnewline
SCFSD &$500$&$0.2581\pm 0.2101$& $0.2786\pm 0.4315$& $1.3933\pm 1.9546$&$ 1.1606\pm 1.3400$\tabularnewline
&$1000$&$0.1537\pm 0.1272$&$ 0.1501\pm 0.1252$&$ 1.6822\pm 2.2284$&$1.2715\pm 1.7100$\\ \hline
&$100$&$ 0.6455 \pm 0.0532$& $ 0.1485\pm 0.1020$& $ 0.6507 \pm 0.2589 $& $2.6174\pm 3.3295$ \tabularnewline
SMSD&$500$&$0.6449\pm 0.0223$& $0.0551\pm 0.0463$&$ 3.7345 \pm 2.2394$& $ 1.3733 \pm 1.3765$\tabularnewline
&$1000$&$ 0.6425 \pm 0.0134$& $ 0.0350\pm 0.0312$& $ 7.7497\pm 1.2857$& $ 0.3811 \pm 0.3846$\\ \hline
\end{tabular}
\end{center}
\end{table}

\begin{figure}
\begin{center}
\includegraphics[width=\columnwidth]{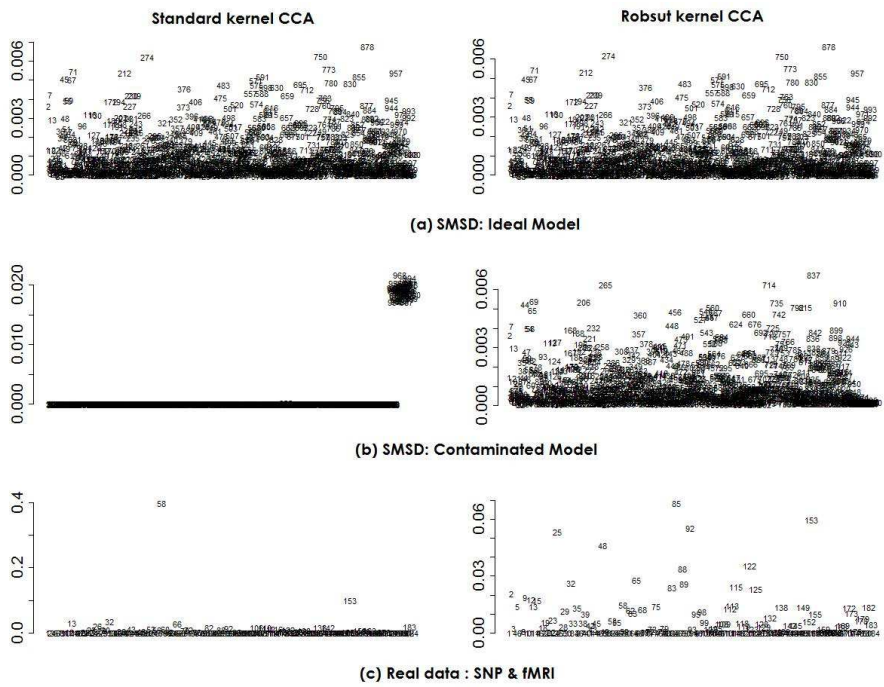}
\caption{Influence  points of standard and robust kernel CCA methods using (a) SMSD ideal model, (b) SMSD contaminated model, (c) Real data: SNP \& fMRI.}
\label{SMDSIFOB}
\end{center}
\end{figure}

Second, we considered a simple graphical display based on the EIF of kernel CCA, the index plots (the subject on the $x$-axis and the influence, $\eta_{\rho}$, on the $y$ axis), to assess the related influence in data fusion regarding EIF based on kernel CCA, $\eta_{\rho}$. To do this, we considered a simulated  data set, SMSD. The index plots of the standard kernel CCA and robust kernel CCA using the SMSD are presented in Figure \ref{SMDSIFOB}. The $1$st and $2$nd rows are for the  ideal and contaminated, and $1$st and $2$nd columns are for the standard kernel CCA (Standard kernel CCA) and robust kernel CCA (Robust kernel CCA), respectively. These plots show that both methods have almost similar results for the ideal data. But for contaminated data, the standard kernel CCA is affected by the contaminated data significantly. We can easily identify the influence of observation using this visualization. On the other hand, the robust kernel CCA has almost similar results for the ideal  and contaminated data. 

Table \ref{tb3:cor} presents the mean and standard deviation for the difference between the training and test of correlation for $10$-fold cross-validation of MGSD and SCFSD simulated data, using standard kernel CCA and robust kernel CCA. From the table, we can conclude that standard kernel CCA is sensitive to the contamination for both data sets. On the other hand, the robust kernel CCA is not only less sensitive to the contaminated data, but also performs better than the standard kernel CCA.

\begin{table}
 \begin{center}
\caption {Mean and standard deviation of  the  differences between the  training and test correlation  in  $10$-fold cross-validation using standard kernel CCA (Standard) and  robust kernel CCA (Robust), respectively.}
\label{tb3:cor}
 \begin{tabular}{llcccccccc} \hline
\rm{Data}&&$\rm{Standard}$&$\rm{Robust}$ \\ \hline
\rm{MGSD}& \rm{ID} &$0.4151\pm 0.210$&$0.3119\pm 0.09140$\\
& \rm{CD} &$0.3673\pm 0.1196$&$0.2609\pm 0.09660$\\
\rm{SCFSD}& \rm{ID} &$0.0002\pm 0.0.0001$&$0.0002\pm 0.0001$\\
& \rm{CD} &$0.0003\pm 0.0002$&$0.0002\pm 0.0001$\\ \hline
\end{tabular}
\end{center}
\end{table}

\subsection{Application to imaging genetics data from MCIC and TCGA}
\label{Sec:apr}
To demonstrate the application of the proposed methods, we used three data sets: the Mind Clinical Imaging Consortium (MCIC) and two data sets from the Cancer Genome Atlas (TCGA) project. The MCIC has collected three types of data: SNPs (723,404 loci), fMRI (51,056 voxels) and DNA methylation (9273 methylation profiles) from 208 subjects including $92$ schizophrenic patients (age: $34\pm 11$, $22$ females) and $116$ (age: $32\pm 11$, $44$ females) healthy controls. Without missing information, the number of subjects is reduced to $183$ ($79$ schizophrenia (SZ) patients and $104$ healthy controls). The detailed information of the MCIC data set is given in \cite{Ashad-16a}. In addition, we consider ovarian serous cystadenocarcinoma (OVSC) and lung squamous cell carcinoma (LUSC) data sets from TCGA data portal. The RNA-Seq gene expression data and methylation profiles are selected from the OVSC and the LUSC patients. After merging the RNA-Seq and methylation data, the number of OVSC patients and LUSC patients are $294$ and $130$, respectively (https://tcga-data.nci.nih.gov/tcga/).  

 To detect influential subjects, we use the EIF of the kernel CC for the standard and robust kernel CCA methods. For robust  kernel CCA, we consider robust kernel CC and kernel CVs as in Theorem \ref{TIFKCCA}.  However, both  standard and robust kernel CCA have identified a similar subject, but robust kernel CCA  is less sensitive than standard kernel CCA. After getting the  influence of the subject, we extracted the outlier subjects of each data set  based on the  `getOutliers' function  of ``extremevalues" R packages.  The outlier subjects of  SNP and fMRI; SNP and Methylation; and fMRI and Methylation are

\[\{7,  31,  36,  41,  55,  58,  60,  72,  80,  92, 140, 150, 162, 165, 168\},\]
\[\{3,  4,  7,  9 ,12, 14 ,51, 55, 56, 58, 60, 61, 62, 66, 80, 88, 90, 93\}\] and \[\{ 6,  15,  39,  41,  58,  59,  69,  72,  83, 107, 116, 132, 133, 134, 140, 148, 162\},\] respectively.  We observed that the  SZ patient number $58$ was common  in  all  cases. In the clinical assessment, this patient has high current   psychosis disorder diagnosis rate ($259.9$). For TCGA data,  outlier patients of OVSC and LUSC are
\[\{6,   8,  11,  20,  25,  29,  31,  60,  94, 121, 159, 172, 175, 199, 231, 236, 264, 287 \} \] and
\[\{15,  37,  39,  54,  57,  58,  59,  71,  72,  81, 120\},\] respectively.

Finally, we investigated the difference between training and testing for correlations using $10$ fold cross-validation. Table \ref{tb:Trc} shows the results of all subjects and all but outlier subjects of MCIC and TCGA data sets using standard kernel CCA and robust kernel CCA. When comparing the subjects with the outliers to the subjects without the outliers the standard kernel CCA method produces drastically different results. Whereas when the robust kernel CCA method is used to compare the two, the results are similar.

\begin{table}
 \begin{center}
\caption {Mean and standard deviation of  the  differences between taring and test correlation of 10 fold cross-validation of MICI and TCGA  data sets using standard kernel CCA (Standard) and  robust kernel CCA (Robust).}
\label{tb:Trc}
 \begin{tabular}{llcccccccc} \hline
\rm{Data}&&&$\rm{Standard}$&$\rm{Robust}$ \\ \hline
&\rm{SNP \& fMRI}& \rm{All} &$0.8107\pm 0.1782$&$0.7867\pm 0.13012$\\
&& \rm{Without outliers} &$0.7361\pm 0.1494$&$0.7348\pm 0.1299$\\
\rm{MCIC}&\rm{SNP \& Methylation}& \rm{All} &$0.7337\pm 0.2000$&$0.7639\pm 0.1433$\\
&& \rm{Without outliers} &$0.6606\pm 0.1772$&$0.7852\pm 0.1776$\\
&\rm{fMRI \&Methylation}& \rm{All} &$0.8424\pm 0.1803$&$0.7842\pm 0.1219$\\
&&\rm{Without outliers} &$0.7479\pm 0.1671$&$0.7617\pm 0.1759$\\\hline
&\rm{OVSC}& \rm{All} &$0.1679\pm 0.0611$&$0.1792\pm 0.0631$\\
\rm{TCGA}&&\rm{Without outliers} &$0.2335\pm 0.0482$&$0.1976\pm 0.0525$\\
&\rm{LISC}& \rm{All} &$0.0779\pm 0.0411$&$0.0713\pm 0.0369$\\
&&\rm{Without outliers} &$0.1349\pm 0.0547$&$0.0987\pm 0.0597$\\\hline
\end{tabular}
\end{center}
\end{table}

\section{Concluding remarks and future research}
\label{sec:conld}
The robust estimator employs a robust loss function instead of a quadratic loss function for the analysis of contaminated data. The robust estimators are  weighted estimators where smaller weights are given  more outlying data points. The weights can be estimated efficiently  using  a KIRLS approach. In terms of accuracy and sensitivity, it is clear that the robust  estimators (e.g., robust kernel CO and robust kernel CCO) perform better than standard estimators (e.g., kernel CO and kernel CCO). We propose the IF of  kernel CCA (kernel CC and kernel CVs) and  robust kernel CCA based on the robust kernel CO and robust kernel CCO.  The proposed IF measures the sensitivity of kernel CCA, which shows that the standard kernel CCA is sensitive to the contamination, but the proposed robust kernel CCA is not. The visualization  method can identify influential (outlier) data in both  synthesized and real imaging genetics data.

While M- estimator based methods are robust with a high breakdown point, finding the theoretical IF of robust kernel CCA is  a future research direction.  Although the focus of this paper is on kernel CCA, we are able to robustify other kernel methods, which must deal with the issue of kernel CO and  kernel CCO. In future work, it would be  interesting to develop robust multiple kernel PCA and robust multiple weighted kernel CCA.

\section*{Appendix A: Proofs}
 We  recall some definitions on  Hilbert spaces. {\em Hilbert-Schmidt  operator} and  {\em Hilbert-Schmidt  norm}  will be used in the proofs of  Lemma \ref{lemma1}, Theorem \ref{thm1} and \ref{thm2}.  

 Let $\mc{H}_1$ and  $\mc{H}_2$ be separable Hilbert spaces. A linear operator  $\mc{T}: \mc{H}_1 \to \mc{H}_2$ is called the {\em Hilbert-Schmidt  operator}  if $\sum_{i}\| \mc{T} \phi_i \|^2 <\infty$ for  an orthonormal basis $\{\phi_i\}_{i\in I}$ of $\mc{H}_1$ with index set $I$.  The sum  $\sum_{i\in I}\|\mc{T} \phi_i \|^2$ does not depend on the orthonormal basis  $\{\phi_i\}_i$.  The  square root of this sum is the  called   {\em Hilbert-Schmidt  norm}, $\|\mc{T}\|_{\rm{HS}}= \sqrt{\sum_{i}\| \mc{T} \phi_i \|^2}$.

\subsection{Proof of Lemma \ref{lemma1}}
 To prove, we need to calculate the G\^{a}teaux differential of $J$.  Let $\Sigma$ and $\mc{T}$ be two  Hilbert-Schmidt  operators. We consider the two cases: $\tilde{\vc{\Phi}}(X_i)\otimes \tilde{\vc{\Phi}} (Y_i)- (\Sigma+\epsilon \mc{T})=0$ and  $\tilde{\vc{\Phi}}(X_i)\otimes \tilde{\vc{\Phi}} (Y_i)- (\Sigma+\epsilon \mc{T})\neq 0$. As in Section \ref{sec:rkme}, $\tilde{\vc{\Phi}}(\cdot)$'s are centered feature maps,    $\zeta (\cdot)$ is a robust loss function, $\zeta^\prime(\cdot)$ is the derivative of $\zeta(\cdot)$, and   $\varphi(t)=\frac{\zeta^\prime(t)}{t}$. 

 Case 1:  $\tilde{\vc{\Phi}}(X_i)\otimes \tilde{\vc{\Phi}} (Y_i)- (\Sigma+\epsilon \mc{T})\neq \vc{0}$
\begin{align}
&\frac{\partial}{\partial\epsilon}\zeta(\|\tilde{\vc{\Phi}} (X_i)\otimes \tilde{\vc{\Phi}} (Y_i)-(\Sigma+\epsilon \mc{T})\|_{\rm{HS}})  \nonumber\\
&=\zeta^\prime (\|\tilde{\vc{\Phi}} (X_i)\otimes \tilde{\vc{\Phi}} (Y_i)-(\Sigma+\epsilon \mc{T})\|_{\rm{HS}})\cdot\frac{\partial}{\partial\epsilon}\|\tilde{\vc{\Phi}} (X_i)\otimes \tilde{\vc{\Phi}} (Y_i)-(\Sigma+\epsilon \mc{T})\|_{\rm{HS}} \nonumber \\
&=\zeta^\prime (\|\tilde{\vc{\Phi}} (X_i)\otimes \tilde{\vc{\Phi}} (Y_i)-(\Sigma+\epsilon \mc{T})\|_{\rm{HS}})\cdot\frac{\partial}{\partial\epsilon} \sqrt{\|\tilde{\vc{\Phi}} (X_i)\otimes \tilde{\vc{\Phi}} (Y_i)-(\Sigma+\epsilon \mc{T})\|^2_{\rm{HS}}} \nonumber \\
&=\zeta^\prime (\|\tilde{\vc{\Phi}} (X_i)\otimes \tilde{\vc{\Phi}} (Y_i)-(\Sigma+\epsilon \mc{T})\|_{\rm{HS}})\cdot\frac{ \frac{\partial}{\partial\epsilon} \|\tilde{\vc{\Phi}} (X_i)\otimes \tilde{\vc{\Phi}} (Y_i)-(\Sigma+\epsilon \mc{T})\|^2_{\rm{HS}}} {2 \sqrt{\varphi (\|\tilde{\vc{\Phi}} (X_i)\otimes \tilde{\vc{\Phi}} (Y_i)-(\Sigma+\epsilon \mc{T})\|^2_{\rm{HS}}}}\nonumber \\
&=\frac{\zeta^\prime (\|\tilde{\vc{\Phi}} (X_i)\otimes \tilde{\vc{\Phi}} (Y_i)-(\Sigma+\epsilon \mc{T})\|_{\rm{HS}})}  {2 \|\tilde{\vc{\Phi}} (X_i)\otimes \tilde{\vc{\Phi}} (Y_i)-(\Sigma+\epsilon \mc{T})\|_{\rm{HS}}} \cdot \frac{\partial}{\partial\epsilon} \Big(\|\tilde{\vc{\Phi}} (X_i)\otimes \tilde{\vc{\Phi}} (Y_i)-\Sigma\|^2_{\rm{HS}}\nonumber \\&\qquad\qquad\qquad- 2 \langle  \tilde{\vc{\Phi}} (X_i)\otimes \tilde{\vc{\Phi}} (Y_i)-\Sigma, \epsilon \mc{T}\rangle_{\rm{HS}} +\epsilon^2 \|\mc{T}\|^2_{\rm{HS}}\Big)\nonumber 
\end{align}

\begin{align}
\label{ep1}
&=\frac{\zeta^\prime (\|\tilde{\vc{\Phi}} (X_i)\otimes \tilde{\vc{\Phi}} (Y_i)-(\Sigma+\epsilon \mc{T})\|_{\rm{HS}})}  {\|\tilde{\vc{\Phi}} (X_i)\otimes \tilde{\vc{\Phi}} (Y_i)-(\Sigma+\epsilon \mc{T})\|_{\rm{HS}}} \cdot  \left(-  \langle  \tilde{\vc{\Phi}} (X_i)\otimes \tilde{\vc{\Phi}} (Y_i)-\Sigma,  \mc{T}\rangle_{\rm{HS}} +\epsilon \|\mc{T}\|^2_{\rm{HS}}\right)\nonumber \\
&=\varphi (\|\tilde{\vc{\Phi}} (X_i)\otimes \tilde{\vc{\Phi}} (Y_i)-(\Sigma+\epsilon \mc{T})\|_{\rm{HS}}) \cdot  \left(-  \langle  \tilde{\vc{\Phi}} (X_i)\otimes \tilde{\vc{\Phi}} (Y_i)-(\Sigma+ \epsilon \mc{T}), \mc{T}\rangle_{\rm{HS}}\right)
\end{align}
Case 2:  $\tilde{\vc{\Phi}}(X_i)\otimes \tilde{\vc{\Phi}} (Y_i)- (\Sigma+\epsilon L)=\vc{0}$
\begin{align}
\label{ep2}
&\frac{\partial}{\partial\epsilon}\zeta(\|\tilde{\vc{\Phi}} (X_i)\otimes \tilde{\vc{\Phi}} (Y_i)-(\Sigma+\epsilon \mc{T})\|_{\rm{HS}})  \nonumber\\
&= \lim_{\delta\to 0} \frac{\zeta(\|\tilde{\vc{\Phi}} (X_i)\otimes \tilde{\vc{\Phi}} (Y_i)-(\Sigma+(\epsilon+\delta) \mc{T})\|_{\rm{HS}}) - \zeta(\|\tilde{\vc{\Phi}} (X_i)\otimes \tilde{\vc{\Phi}} (Y_i)-(\Sigma+\epsilon \mc{T})\|_{\rm{HS}}) }{\delta}  \nonumber\\
&= \lim_{\delta\to 0}\frac{ \zeta (\|\delta \mc{T}\|_{\rm{HS}})- \zeta (0)}{\delta}.  
 \end{align}
For $\mc{T}=\vc{0}$ and $\mc{T}\neq \vc{0}$  the above equation is  equal to   $\lim_{\delta\to 0} \frac{\zeta (0)}{\delta}$ and $\frac{\zeta (\|\delta \mc{T}\|_{\rm{HS}})}{\delta}\cdot \| \mc{T}\|_{\rm{HS}}$, receptively. Using the assumption (i) we have  $\frac{\partial}{\partial\epsilon}\zeta(\|\tilde{\vc{\Phi}} (X_i)\otimes \tilde{\vc{\Phi}} (Y_i)-(\Sigma+\epsilon \mc{T})\|_{\rm{HS}})=0$. Since  $\tilde{\vc{\Phi}}(X_i)\otimes \tilde{\vc{\Phi}} (Y_i)- (\Sigma+\epsilon \mc{T})=\vc{0}$ and $\varphi(0)$ is well-defined by the assumption (ii). Combining Eq. (\ref{ep1}) and   Eq. (\ref{ep2}) we get 
\begin{align}
 \frac{\partial}{\partial\epsilon}\zeta(\|\tilde{\vc{\Phi}} (X_i)\otimes \tilde{\vc{\Phi}} (Y_i)-(\Sigma+\epsilon \mc{T})\|_{\rm{HS}} = \nonumber \\\varphi (\|\tilde{\vc{\Phi}} (X_i)\otimes \tilde{\vc{\Phi}} (Y_i)-(\Sigma+\epsilon \mc{T})\|_{\rm{HS}}) \cdot  \left(-  \langle  \tilde{\vc{\Phi}} (X_i)\otimes \tilde{\vc{\Phi}} (Y_i)-(\Sigma+ \epsilon \mc{T}), \mc{T}\rangle_{\rm{HS}}\right).
 \end{align}
 Now it is clear that for any $\Sigma, \mc{T} \in \rm{HS}$, 
  \begin{align}
&\frac{\partial}{\partial\epsilon}\zeta(\|\tilde{\vc{\Phi}} (X_i)\otimes \tilde{\vc{\Phi}} (Y_i)-(\Sigma+\epsilon \mc{T})\|_{\rm{HS}}  \nonumber\\
&=\varphi (\|\tilde{\vc{\Phi}} (X_i)\otimes \tilde{\vc{\Phi}} (Y_i)-(\Sigma+\epsilon \mc{T})\|_{\rm{HS}}) \cdot  \left(-  \langle  \tilde{\vc{\Phi}} (X_i)\otimes \tilde{\vc{\Phi}} (Y_i)-(\Sigma+ \epsilon \mc{T}), \mc{T}\rangle_{\rm{HS}}\right)
\end{align}

Therefore, 
\begin{align}
&\delta J(\Sigma; \mc{T}) = \frac{\partial}{\partial\epsilon}J(\Sigma\epsilon \mc{T})\big|_{\epsilon=0} \nonumber\\
&=\frac{\partial}{\partial\epsilon} \left(\frac{1}{n}\sum_{i=1}^n\zeta (\|\tilde{\vc{\Phi}} (X_i)\otimes \tilde{\vc{\Phi}} (Y_i)-(\Sigma+\epsilon \mc{T})\|_{\rm{HS}})\right)\big|_{\epsilon=0} \nonumber \\
&=\frac{1}{n}\sum_{i=1}^n\frac{\partial}{\partial\epsilon} \zeta (\|\tilde{\vc{\Phi}} (X_i)\otimes \tilde{\vc{\Phi}} (Y_i)-(\Sigma+\epsilon \mc{T})\|_{\rm{HS}})\big|_{\epsilon=0} \nonumber \\
&=\frac{1}{n}\sum_{i=1}^n \varphi (\|\tilde{\vc{\Phi}} (X_i)\otimes \tilde{\vc{\Phi}} (Y_i)-(\Sigma+\epsilon \mc{T})\|_{\rm{HS}}) \cdot  \left(-  \langle  \tilde{\vc{\Phi}} (X_i)\otimes \tilde{\vc{\Phi}} (Y_i)-(\Sigma+ \epsilon L), \mc{T}\rangle_{\rm{HS}}\right) \big|_{\epsilon=0}  \nonumber \\
&=-\frac{1}{n}\sum_{i=1}^n \varphi (\|\tilde{\vc{\Phi}} (X_i)\otimes \tilde{\vc{\Phi}} (Y_i)-\Sigma\|_{\rm{HS}}) \cdot \langle\tilde{\vc{\Phi}} (X_i)\otimes \tilde{\vc{\Phi}} (Y_i)-\Sigma, \mc{T}\rangle_{\rm{HS}} \nonumber \\
&=- \left \langle \frac{1}{n}\sum_{i=1}^n \varphi (\|\tilde{\vc{\Phi}} (X_i)\otimes \tilde{\vc{\Phi}} (Y_i)-\Sigma\|_{\rm{HS}}) \cdot (\tilde{\vc{\Phi}} (X_i)\otimes \tilde{\vc{\Phi}} (Y_i)-\Sigma), \mc{T} \right\rangle_{\rm{HS}} \nonumber \\
&=- \left \langle S(\Sigma), \mc{T} \right\rangle_{\rm{HS}}
\end{align}
The necessary condition for $\Sigma$ to be a minimizer of $J$, i.e., $\Sigma= \widehat{\Sigma}_{RXY}$, is that $\delta J(\Sigma; \mc{T})=0\, \forall\, \mc{T}\in \rm{HS}$, which leads to $S(\Sigma)=\vc{0}.$

\subsection{Proof of Theorem \ref{thm2}}
Using Lemma \ref{lemma1} we have
$\varphi (\|\tilde{\vc{\Phi}} (X_i)\otimes \tilde{\vc{\Phi}} (Y_i)-\widehat{\Sigma}_{RXY})\|_{\rm{HS}}) \cdot (\tilde{\vc{\Phi}} (X_i)\otimes \tilde{\vc{\Phi}} (Y_i)-\widehat{\Sigma}_{RXY}))=\vc{0}$
By solving  $\widehat{\Sigma}_{RXY}$, we get  $\widehat{\Sigma}_{RXY}=\sum_{i=1}^nw_i \tilde{\vc{\Phi}} (X_i)\otimes \tilde{\vc{\Phi}} (Y_i)$, where
\[w_i= \left ( \varphi (\|\tilde{\vc{\Phi}} (X_i)\otimes \tilde{\vc{\Phi}} (Y_i)-\widehat{\Sigma}_{RXY})\|_{\rm{HS}}) \right)^{-1} \cdot \varphi (\|\tilde{\vc{\Phi}} (X_i)\otimes \tilde{\vc{\Phi}} (Y_i)-\widehat{\Sigma}_{RXY})\|_{\rm{HS}}).\] 
Since the function $\zeta$ is non-deceasing, $w_i\geq 0$ and $\sum_{i=1}^nw_i=1$.

\subsection{Proof of Theorem \ref{thm3}}
We will prove this theorem in three steps; (i)  the monotone decreasing property of $J(\Sigma^{(h)})$, (ii) every limit point $\Sigma^*$ of $\{ \Sigma^{(h)}\}_{h=1}^\infty \in U$, and (iii) by contradiction . We define a function
\[ p(t; c)= \zeta(c)- \frac{1}{2} t\zeta^\prime(c)+\frac{1}{2} \varphi(c) t^2,\]
where $c$ is a real number. As shown in \citep{Huber-09}, for  non increasing $\varphi$,  the function $p$ is a surrogate function of $\zeta$ with the following two properties
\begin{eqnarray}
\label{th5e1}
p(c; c)= \zeta(c) 
\end{eqnarray}
and
\begin{eqnarray}
\label{th5e2}
p(t; c)\geq \zeta(t)\, \quad \forall t.
\end{eqnarray}
Now we define a bivariate function

\[ \mc{Q}(\Sigma; \Sigma^{(h)}) = \frac{1}{n}\sum_{i=1}^n p(\|\tilde{\vc{\Phi}}(X_i))\times \tilde{\vc{\Phi}}(Y_i) -\Sigma\|_{\rm{HS}};  \|\tilde{\vc{\Phi}}(X_i)\times \tilde{\vc{\Phi}}(Y_i)-\Sigma^{(h)}\|_{\rm{HS}}),\]

which is a continuous function in both arguments because  both $\zeta^\prime$ and $\varphi$ are continuous functions.

Step (i): using Eq. (\ref{th5e1}) and  Eq. (\ref{th5e2}), we have

\begin{align}
\label{th5e3}
\mc{Q}(\Sigma^{(h)}; \Sigma^{(h)}) &= \frac{1}{n}\sum_{i=1}^n p(\|\tilde{\vc{\Phi}}(X_i)\times \tilde{\vc{\Phi}}(Y_i) -\Sigma^{(h)}\|_{\rm{HS}};  \|\tilde{\vc{\Phi}}(X)_i\times \tilde{\vc{\Phi}}(Y_i) -\Sigma^{(h)}\|_{\rm{HS}})  \nonumber \\
&=\frac{1}{n}\sum_{i=1}^n \rho(\|\tilde{\vc{\Phi}}(X_i)\times \tilde{\vc{\Phi}}(Y_i) -\Sigma^{(h)}\|_{\rm{HS}};  \|\tilde{\vc{\Phi}}(X_i)\times \tilde{\vc{\Phi}}(Y_i) -\Sigma^{(h)})  \nonumber \\
&= J(\Sigma^{(h)})
\end{align}
 and 
\begin{align}
\label{th5e4}
\mc{Q}(\Sigma; \Sigma^{(h)}) &= \frac{1}{n}\sum_{i=1}^n p(\|\tilde{\vc{\Phi}}(X_i)\times \tilde{\vc{\Phi}}(Y_i) -\Sigma\|_{\rm{HS}};  \|\tilde{\vc{\Phi}}(X_i)\times \tilde{\vc{\Phi}}(Y_i) -\Sigma^{(h)}\|_{\rm{HS}})  \nonumber \\
&\geq \frac{1}{n}\sum_{i=1}^n \rho(\|\tilde{\vc{\Phi}}(X_i)\times \tilde{\vc{\Phi}}(Y_i) -\Sigma\|_{\rm{HS}})  \nonumber \\
&= J(\Sigma), \, \forall \, \Sigma \in {\rm{HS}}.
\end{align}

Now,
\begin{align}
\label{th5e5}
\Sigma^{(h+1)}&= \sum_{i=1}^n w_i^{(h)}\tilde{\vc{\Phi}}(X_i)\otimes\tilde{\vc{\Phi}}(Y_i) \nonumber \\
&=  \sum_{i=1}^n \frac{\varphi(\|\tilde{\vc{\Phi}}(X_i)\otimes \tilde{\vc{\Phi}}(Y_i) - \Sigma^{(h)}\|_{\mc{H}_X\otimes\mc{H}_Y})}{\sum_{b=1}^n\varphi(\| \tilde{\vc{\Phi}}(X_b)\otimes\tilde{\vc{\Phi}}(Y_b)- \Sigma^{(h)}\|_{\rm{HS}}}\tilde{\vc{\Phi}}(X_i)\otimes\tilde{\vc{\Phi}}(Y_i) \nonumber\\
&= \argmin_{\Sigma\in{\rm{HS}}}  \sum_{i=1}^n  \varphi(\|\tilde{\vc{\Phi}}(X_i)\otimes \tilde{\vc{\Phi}}(Y_i) - \Sigma^{(h)}\|_{\rm{HS}})\cdot \|\tilde{\vc{\Phi}}(X_i)\otimes \tilde{\vc{\Phi}}(Y_i) - \Sigma\|^2_{\rm{HS}}\nonumber\\
&= \argmin_{\Sigma\in{\rm{HS}}} \mc{Q}(\Sigma; \Sigma^{(h)})
\end{align}

 From  Eq. (\ref{th5e3}),  Eq. (\ref{th5e4}), and  Eq. (\ref{th5e5}), we have
\[J(\Sigma^{(h)}) = \mc{Q}(\Sigma^{(h)};\Sigma^{(h)} ) \geq  \mc{Q}(\Sigma^{(h+1)};\Sigma^{(h)}) \geq J(\Sigma^{(h+1)}).\]

Thus, it is proved that  $J(\Sigma^{(h)})$ monotonically decreases at every iteration. In addition, since   $J(\Sigma^{(h)})\geq 0$, for any $h \geq 1$ (the sequence is bounded below at 0),  it converses.

Step (ii): as in \citep{Kim-12}, it is clear that $\{ \Sigma^{(h)}\}_{h=1}^\infty$   has a convergent subsequence   $\{ \Sigma^{(h_\ell)}\}_{\ell=1}^\infty$.  Let $\Sigma^*$  be the limit of $\{ \Sigma^{(h_\ell)}\}_{\ell=1}^\infty$.  Using  Eq. (\ref{th5e3}),  Eq. (\ref{th5e4}),  Eq. (\ref{th5e5}), and the monotone decreasing property of $J(\Sigma^{(h)})$,  we have also get

\begin{align}
\mc{Q}(\Sigma^{(h_{\ell+1})}; \Sigma^{(h_{\ell+1})}) &=J(\Sigma^{(h_{\ell+1})})\nonumber\\
&\leq  J(\Sigma^{(h_{\ell+1})})\nonumber\\
&\leq  \mc{Q}(\Sigma^{(h_{\ell+1})}; \Sigma^{(h_{\ell})}) \nonumber\\
&\leq  \mc{Q}(\Sigma; \Sigma^{(h_{\ell})}), \, \forall \Sigma\in \rm{HS}. \nonumber 
\end{align}
Taking the limit on  both sides of the above inequality, we have
\[ \mc{Q}(\Sigma^*; \Sigma^* ) \leq  D(\Sigma; \Sigma^*), \, \, \forall \Sigma\in \rm{HS}.\]

Therefore,
\begin{align}
\Sigma^*&= \argmin_{\Sigma\in\rm{HS}} \mc{Q}(\Sigma; \Sigma^*)\nonumber\\
&= \frac{\varphi(\|\tilde{\vc{\Phi}}(X_i)\otimes \tilde{\vc{\Phi}}(Y_i) - \Sigma^*\|_{\rm{HS}})}{\sum_{b=1}^n\varphi(\| \tilde{\vc{\Phi}}(X_b)\otimes\tilde{\vc{\Phi}}(Y_b)- \Sigma^*\|_{\rm{HS}})}\tilde{\vc{\Phi}}(X_i)\otimes\tilde{\vc{\Phi}}(Y_i) 
\end{align}

and thus
\[ \varphi(\|\tilde{\vc{\Phi}}(X_i)\otimes \tilde{\vc{\Phi}}(Y_i) - \Sigma^*\|_{\rm{HS}})\cdot (\tilde{\vc{\Phi}}(X_i)\otimes \tilde{\vc{\Phi}}(Y_i) - \Sigma^*) =\vc{0}.\]
This implies $\Sigma^*\in U.$

Step (iii): suppose $\inf_{\Sigma\in U }\| \Sigma^{(h)}-  \Sigma\|_{\rm{HS}}$ does not tend to $0$. Then there exists  $\epsilon > 0$ such that  $\forall I\in \mb{N}, \exists h > I$ with $\inf_{\Sigma\in U }\| \Sigma^{(h)}-  \Sigma\|_{\rm{HS}} \geq \epsilon$.  Thus  we are able to regard a  increasing sequence of indices  such that the  $\inf_{\Sigma\in U }\| \Sigma^{(h_\ell)}-  \Sigma\|_{\rm{HS}} \geq \epsilon$ for all $\ell= 1, 2,\cdots \cdots$. Since  $\Sigma^{(h_\ell)}$ lies in the compact subset of  $\rm{HS}$, it has a subsequence converging to some $	\Sigma^\dagger$, and we can choose $j$ such that $\|\Sigma^{(h_j)} -\Sigma^\dagger\|_{\rm{HS}}  < \epsilon/2$. Since  $\Sigma^\dagger$ is also a limit point of $\{ \Sigma^{(h)}\}_{h=1}^\infty$, $\Sigma^\dagger \in U$. This is  a contradiction because
\[ \epsilon \leq \inf_{\Sigma\in U }\| \Sigma^{(h)}-  \Sigma\|_{\rm{HS}}\leq \| \Sigma^{(h)}-  \Sigma^\dagger\|_{\rm{HS}} < \epsilon/2.\]

\subsection{Proof of Theorem \ref{TIFKCCA}}
We present the derivation of the IF of standard kernel CCA in detail. Recall the generalized eigenvalue problem  in  Eq. (\ref{ckcca4}). We   can  formulate this problem as a simple eigenvalue problem. Using the  {\it j}-th eigenfunction of  the first equation  of Eq. (\ref{ckcca4})  we have 

\begin{eqnarray}
\label{ckcca5a}
(\Sigma_{XX}^{- \frac{1}{2}} \Sigma_{XY} \Sigma_{YY}^{-1}\Sigma_{YX}\Sigma_{XX}^{- \frac{1}{2}} - \rho_j^2I) \Sigma_{XX}^{\frac{1}{2}}f_{jX}&=&0 \nonumber \\
\Rightarrow (\Sigma_{XX}^{-\frac{1}{2}} \Sigma_{XY} \Sigma_{YY}^{-1}\Sigma_{YX}\Sigma_{XX}^{- \frac{1}{2}} - \rho_j^2I)e_{jX}& = &0
\end{eqnarray}
where $e_{jX} = \Sigma_{XX}^{\frac{1}{2}}f_{jX}$.

 To establish the IF of kernel CCA, we  convert the generalized eigenvalue problem of kernel CCA  into a simple eigenvalue problem. Henceforth, we can use the results  such as  the IF of  linear PCA analysis \citep{Tanaka-88, Tanaka-89},  the IF of linear CCA \citep{Romanazii-92}  and the IF of  kernel PCA  (finite dimension  and   infinite dimension) \citep {Huang-KPCA}, respectively. To do this, first  we calculate  the IF of  $ \Sigma_{XX}^{- \frac{1}{2}} \Sigma_{XY} \Sigma_{YY}^{-1}\Sigma_{YX}\Sigma_{XX}^{- \frac{1}{2}}$,  henceforth,  the IF of $\Sigma_{YY}^{-1},  \Sigma_{XX}^{\frac{1}{2}}$ and $\Sigma_{XY}$.

 Using simple algebra (as in Section{\ref{sec:ifK}),  we have the IF for the   following  operators  at point $Z^\prime= (X^\prime, Y^\prime)$:
\begin{align}
&\rm{IF}(\cdot, X^\prime, \Sigma_{XX})= (k_X(\cdot, X^\prime)-\mc{M}_X)  \otimes(k_X(\cdot, X^\prime)-\mc{M}_X)- \Sigma_{XX}, \nonumber\\
&\rm{IF}(\cdot, Y^\prime, \Sigma_{YY})= (k_Y(\cdot, Y^\prime)-\mc{M}_Y) \otimes  (k_Y(\cdot, Y^\prime)-\mc{M}_Y)- \Sigma_{YY},\nonumber\\
&\rm{IF}(\cdot, Z^\prime, \Sigma_{XY})= (k_X(\cdot, X^\prime) -\mc{M}_X)\otimes  (k_Y(\cdot, Y^\prime)-\mc{M}_Y)- \Sigma_{XY} \, \rm{and} \nonumber\\
&\rm{IF}(\cdot, Z^\prime, 
\Sigma_{XX}^{-\frac{1}{2}} )= \frac{1}{2} [\Sigma_{XX}^{-\frac{1}{2}}- \Sigma_{XX}^{-\frac{1}{2}}(k_X(\cdot, X^\prime)-\mc{M}_X)\otimes  (k_X(\cdot, X^\prime)-\mc{M}_X) \Sigma_{XX}^{-\frac{1}{2}}]. \nonumber
\label{IFIVCOVa}
\end{align}
For simplicity, let us define $\tilde{k}_X (\cdot, X^\prime):= k_X(\cdot, X^\prime)-\mc{M}_X, \, \tilde{k}_Y (\cdot, \vc{y}^\prime):= k_Y(\cdot,  Y^\prime) - \mc{M}_Y$. Also define $\mb{A}:= \Sigma_{XY} \Sigma_{YY}^{-1}\Sigma_{YX}, \, \mb{B}:= \Sigma_{XX}^{- \frac{1}{2}}\mb{A}\Sigma_{XX}^{- \frac{1}{2}}$, and $\mb{L}= \Sigma_{XX}^{- \frac{1}{2}}(\Sigma_{XX}^{- \frac{1}{2}} \Sigma_{XY} \Sigma_{YY}^{-1} \Sigma_{YX}\Sigma_{XX}^{- \frac{1}{2}}-\rho^2\vc{I})^{-1}\Sigma_{XX}^{- \frac{1}{2}}$. Now
\begin{align}
&\rm{IF} (\cdot, Z^\prime, \mb{A})= \rm{IF}(X^\prime, Y^\prime, \Sigma_{XY}) \Sigma^{-1}_{YY}\Sigma_{YX} + \Sigma_{XY}  \rm{IF}(X^\prime, Y^\prime, \Sigma_{YY}^{-1})\Sigma_{YX}+ \Sigma_{XY} \Sigma_{YY}^{-1}\rm{IF}(X^\prime, Y^\prime, \Sigma_{XY}) \nonumber\\
&= \left[\tilde{k}_X (\cdot, X^\prime)\otimes \tilde{k}_Y (\cdot, Y^\prime)- \Sigma_{XY}\right]\Sigma_{YY}^{-1}\Sigma_{YX}
+ \Sigma_{XY}\left[\Sigma_{YY}^{-1}-\Sigma_{YY}^{-1}\tilde{k}_Y (\cdot, Y^\prime)\otimes \tilde{k}_Y (\cdot, Y^\prime) \Sigma_{YY}^{-1}\right] \Sigma_{YX}\nonumber\\
&\qquad \qquad\qquad \qquad\qquad \qquad\qquad \qquad+ \Sigma_{XY}\Sigma_{YY}^{-1}\left[\tilde{k}_X (\cdot, X^\prime)\otimes \tilde{k}_Y (\cdot, Y^\prime)- \Sigma_{YX}\right]\nonumber\\
&=2 \Sigma_{XY}\Sigma^{-1}_{YY}\left[\tilde{k}_X(\cdot, X^\prime)\otimes  \tilde{k}_Y(\cdot, Y^\prime)- \Sigma_{XY}\right]
+ \Sigma_{XY}\left[\Sigma_{YY}^{-1}-\Sigma_{YY}^{-1}\tilde{k}_Y (\cdot, Y^\prime)\otimes \tilde{k}_Y (\cdot, Y^\prime) \Sigma_{YY}^{-1}\right] \Sigma_{YX}  \nonumber
\end{align}
Then,
\begin{multline}
\Sigma_{XX}^{-\frac{1}{2}} \rm{IF}(Z^\prime, \mb{A}) \Sigma_{XX}^{-\frac{1}{2}}= 2  \Sigma_{XX}^{-\frac{1}{2}}\Sigma_{XY}\Sigma^{-1}_{YY}[ \tilde{k}_X (\cdot, X^\prime)\otimes \tilde{k}_Y (\cdot, Y^\prime)- \Sigma_{XY}] \Sigma_{XX}^{-\frac{1}{2}}\nonumber\\
+\Sigma_{XX}^{-\frac{1}{2}}\Sigma_{XY}[\Sigma_{YY}^{-1}-\Sigma_{YY}^{-1}[\tilde{k}_Y (\cdot, Y^\prime)\otimes \tilde{k}_Y (\cdot, Y^\prime)] \Sigma_{YY}^{-1}] \Sigma_{YX}\Sigma_{XX}^{-\frac{1}{2}} \nonumber
\end{multline}
and
\begin{multline}
 \rm{IF}(X^\prime, \Sigma_{XX}^{-\frac{1}{2}})\mb{A} \Sigma_{XX}^{-\frac{1}{2}}+ \Sigma_{XX}^{-\frac{1}{2}}\mb{A} \rm{IF}(X^\prime, \Sigma_{XX}^{-\frac{1}{2}})
= \nonumber\\2 \rm{IF}(X^\prime, \Sigma_{XX}^{-\frac{1}{2}})\mb{A} \Sigma_{XX}^{-\frac{1}{2}}
= [\Sigma_{XX}^{-\frac{1}{2}}- \Sigma_{XX}^{-\frac{1}{2}}\tilde{k}_X (\cdot, X^\prime)\otimes \tilde{k}_X (\cdot, X^\prime) \Sigma_{XX}^{-\frac{1}{2}}] \mb{A} \Sigma_{XX}^{-\frac{1}{2}}.\nonumber
\end{multline}
The  influence of $\mb{B}$ is then given by
\begin{align}
&\rm{IF} (X^\prime, Y^\prime,\mb{B})= 2\rm{IF} (X^\prime, Y^\prime, \Sigma_{XX}^{-\frac{1}{2}}) \Sigma_{XY} \Sigma_{YY}^{-1}\Sigma_{XY} \Sigma_{XX}^{-\frac{1}{2}}+ \Sigma_{XX}^{-\frac{1}{2}} \rm{IF}(X^\prime, Y^\prime, \mb{A}) \Sigma_{XX}^{-\frac{1}{2}} \nonumber\\
&=  [\Sigma_{XX}^{-\frac{1}{2}}- \Sigma_{XX}^{-\frac{1}{2}}\tilde{k}_X (\cdot, X^\prime)\otimes \tilde{k}_X (\cdot, X^\prime)\Sigma_{XX}^{-\frac{1}{2}}] \Sigma_{XY} \Sigma_{YY}^{-1}\Sigma_{XY}\Sigma_{XX}^{-\frac{1}{2}} \nonumber\\
& \qquad \qquad \qquad\qquad + 2  \Sigma_{XX}^{-\frac{1}{2}}\Sigma_{XY}\Sigma^{-1}_{YY}[ \tilde{k}_X (\cdot, X^\prime)\otimes \tilde{k}_Y (\cdot, Y^\prime) - \Sigma_{XY}] \Sigma_{XX}^{-\frac{1}{2}}\nonumber\\
&\qquad\qquad\qquad\qquad+  \Sigma_{XX}^{-\frac{1}{2}}\Sigma_{XY}[\Sigma_{YY}^{-1} - \Sigma_{YY}^{-1}\tilde{k}_Y (\cdot, Y^\prime)\otimes \tilde{k}_Y (\cdot, Y^\prime) \Sigma_{YY}^{-1}] \Sigma_{YX}\Sigma_{XX}^{-\frac{1}{2}}\nonumber \\
&= - \Sigma_{XX}^{-\frac{1}{2}}\tilde{k}_X (\cdot, X^\prime)\otimes \tilde{k}_X (\cdot, X^\prime) \Sigma_{XX}^{-\frac{1}{2}} \Sigma_{XY} \Sigma_{YY}^{-1}\Sigma_{XY}\Sigma_{XX}^{-\frac{1}{2}} \nonumber \\
& \qquad \qquad \qquad\qquad+ 2  \Sigma_{XX}^{-\frac{1}{2}}\Sigma_{XY}\Sigma^{-1}_{YY} \tilde{k}_X (\cdot, X^\prime)\otimes \tilde{k}_Y (\cdot, X^\prime) \Sigma_{XX}^{-\frac{1}{2}}\nonumber \\
&\qquad\qquad \qquad\qquad-  \Sigma_{XX}^{-\frac{1}{2}}\Sigma_{XY}\Sigma_{YY}^{-1}\tilde{k}_Y (\cdot, Y^\prime)\otimes \tilde{k}_Y (\cdot, Y^\prime) \Sigma_{YY}^{-1} \Sigma_{YX}\Sigma_{XX}^{-\frac{1}{2}}
\end{align}
To define the IF of kernel CC ($\rho_j^2$) and  kernel CVs ($f_X(X)$ and $f_Y(Y)$),  we convert a generalized eigenvalue problem and use the Lemma 1 of \cite{Huang-KPCA} and Lemma 2 of  \cite{Tanaka-89}.  Then the  IF of kernel $\rho_j^2$ is defined as follows

\begin{align}
&\rm{IF} (Z^\prime,\rho_j^2)= \langle e_{jX}, \rm{IF} (Z^\prime, \mb{B}) e_{jX}\rangle_{\mc{H}_X\otimes \mc{H}_Y}\nonumber\\
&=-\langle e_{jX}, \Sigma_{XX}^{-1}\tilde{k}_X (\cdot, X^\prime)\otimes \tilde{k}_X (\cdot, X^\prime) \Sigma_{XX}^{-\frac{1}{2}} \Sigma_{XY} \Sigma_{YY}^{-1}\Sigma_{XY}\Sigma_{XX}^{-\frac{1}{2}} e_{jX}\rangle_{\mc{H}_X\otimes \mc{H}_X}\nonumber\\
& \qquad \qquad \qquad\qquad+\ 2\langle  e_{jX}, \Sigma_{XX}^{-\frac{1}{2}}\Sigma_{XY}\Sigma^{-1}_{YY}\tilde{k}_X (\cdot, X^\prime)\otimes \tilde{k}_Y (\cdot, Y^\prime) \Sigma_{XX}^{-\frac{1}{2}} e_{jX}\rangle_{\mc{H}_X\otimes \mc{H}_Y}\nonumber\\
&\qquad \qquad \qquad\qquad-\langle e_{jX}^T,  \Sigma_{XX}^{-\frac{1}{2}}\Sigma_{XY}\Sigma_{YY}\tilde{k}_Y (\cdot, Y^\prime)\otimes \tilde{k}_Y (\cdot, Y^\prime)\Sigma_{YX}\Sigma_{XX}^{-\frac{1}{2}} e_{jX}\rangle_{\mc{H}_Y\otimes \mc{H}_Y}\nonumber\\
&=-\langle e_{jX}, \Sigma_{XX}^{-1}\tilde{k}_X (\cdot, X^\prime)\otimes \tilde{k}_X (\cdot, X^\prime) \Sigma_{XX}^{-\frac{1}{2}} \Sigma_{XY} \Sigma_{YY}^{-1}\Sigma_{XY}\Sigma_{XX}^{-\frac{1}{2}} e_{jX}\rangle_{\mc{H}_X\otimes \mc{H}_X}\nonumber\\
& \qquad \qquad \qquad\qquad +2 \langle  e_{jX}^T, \Sigma_{XX}^{-\frac{1}{2}}\Sigma_{XY}\Sigma^{-1}_{YY}\tilde{k}_X (\cdot, X^\prime)\otimes \tilde{k}_Y (\cdot, Y^\prime) \Sigma_{XX}^{-\frac{1}{2}} e_{jX}\rangle_{\mc{H}_X\otimes \mc{H}_Y}\nonumber\\
& \qquad \qquad \qquad\qquad -\langle e_{jX}, \Sigma_{XX}^{-\frac{1}{2}}\Sigma_{XY}[\Sigma_{YY}\tilde{k}_Y (\cdot, Y^\prime)\otimes \tilde{k}_Y (\cdot, Y^\prime) \Sigma_{YY}^{-1}] \Sigma_{YX}\Sigma_{XX}^{-\frac{1}{2}} e_{jX}\rangle_{\mc{H}_Y\otimes \mc{H}_Y}
\label{IFKCCA1a}
\end{align}

For simplicity, Eq. (\ref{IFKCCA1a}) can calculate in parts.  The first part is derived as
\begin{align}
&\langle e_{jX}, \Sigma_{XX}^{-1}\tilde{k}_X (\cdot, X^\prime)\otimes \tilde{k}_X (\cdot, X^\prime)  \Sigma_{XX}^{-\frac{1}{2}} \Sigma_{XY} \Sigma_{YY}^{-1}\Sigma_{XY}\Sigma_{XX}^{-\frac{1}{2}} e_{jX}\rangle_{\mc{H}_X\otimes \mc{H}_X} \nonumber\\
&= \langle \Sigma_{XX}^{-\frac{1}{2}}e_{jX},\tilde{k}_X (\cdot, X^\prime)\otimes \tilde{k}_X (\cdot, X^\prime)  \Sigma_{XX}^{-\frac{1}{2}} \Sigma_{XY} \Sigma_{YY}^{-1}\Sigma_{XY}\Sigma_{XX}^{-\frac{1}{2}}\Sigma_{XX}^{-\frac{1}{2}} e_{jX}\rangle_{\mc{H}_X\otimes \mc{H}_X} \nonumber\\
&= \langle f_{jX}, \tilde{k}_X(\cdot, X^\prime)\rangle_{\mc{H}_X} \langle \tilde{k}_X(\cdot, X^\prime), \Sigma_{XX}^{-\frac{1}{2}} \Sigma_{XY} \Sigma_{YY}^{-1}\Sigma_{XY}\Sigma_{XX}^{-\frac{1}{2}} f_{jX}\rangle_{\mc{H}_X}\nonumber\\
&= \rho_j^2 \tilde{f}_{jX}^2(X^\prime).
\label{IFKCCA2a}
\end{align}
In the last equality, we used   Eq. (\ref{ckcca5a}). The 2nd part of   Eq. (\ref{IFKCCA1a}) is derived as
\begin{align}
&\langle f_{jX}, \Sigma_{XX}^{-\frac{1}{2}}\Sigma_{XY}\Sigma^{-1}_{YY}[ \tilde{k}_X (\cdot, X^\prime)\otimes \tilde{k}_Y (\cdot, Y^\prime)] \Sigma_{XX}^{-\frac{1}{2}} f_{jX}\rangle_{\mc{H}_X\otimes \mc{H}_Y} \nonumber\\
&= \langle \Sigma_{XX}^{-\frac{1}{2}} e_{jX}, \tilde{k}_X (\cdot, X^\prime)\otimes \tilde{k}_Y (\cdot, Y^\prime)\Sigma_{XY}\Sigma^{-1}_{YY} \Sigma_{XX}^{-\frac{1}{2}} f_{jX}\rangle_{\mc{H}_X\otimes \mc{H}_Y} \nonumber\\
&=\langle f_{jX},\tilde{k}_X (\cdot, X^\prime)\otimes \tilde{k}_Y (\cdot, Y^\prime) \Sigma_{XY}\Sigma^{-1}_{YY}  f_{jX}\rangle_{\mc{H}_X\otimes \mc{H}_Y} \nonumber\\
&=  \rho_j \langle f_{jX}, \tilde{k}_X (\cdot, X^\prime)\rangle_{\mc{H}_X} \langle \tilde{k}_Y (\cdot, Y^\prime),  f_{jY}\rangle_{\mc{H}_Y}
 \nonumber\\
&=  \rho_j \tilde{f}_{jX}( X^\prime) \tilde{f}_{jY}(Y^\prime).
\label{IFKCCA3a}
\end{align}
In the last second equality, we used   Eq. (\ref{ckcca4}). Similarly we can write the 3rd term as
\begin{align}
\label{IFKCCA4a}
\langle e_{jX},  \Sigma_{XX}^{-\frac{1}{2}}\Sigma_{XY}[\Sigma_{YY}\tilde{k}_Y (\cdot, Y^\prime)\otimes \tilde{k}_Y (\cdot, Y^\prime)  \Sigma_{YY}^{-1}] \Sigma_{YX}\Sigma_{XX}^{-\frac{1}{2}} e_{jX}\rangle_{\mc{H}_Y\otimes \mc{H}_Y} =   \rho_j^2 \tilde{f}_{jY}^2(Y^\prime)
\end{align}
where  $\tilde{f}_{jX}(X^\prime)= \langle f_{jX}, \tilde{k}_X (\cdot, X^\prime) \rangle$ and similar  for $\tilde{f}_{jY}$.
Therefore, substituting Eq. (\ref{IFKCCA2a}), (\ref{IFKCCA3a})  and (\ref{IFKCCA4a}) into Eq. (\ref{IFKCCA1a}), the IF  of kernel CC is given by
\begin{align}
\rm{IF} (X^\prime, Y^\prime,\rho_j)= - \rho_j^2 \tilde{f}_{jX}^2(Y^\prime)+2 \rho_j\tilde{f}_{jX}(X^\prime) \tilde{f}_{jY}(Y^\prime)  -\rho_j^2 \tilde{f}_{jY}^2(Y^\prime)
\end{align}
Now we derived the IF of  kernel CVs. To this end, first we need to derive
\begin{align}
\rm{IF} (X^\prime, f_{jx})= \rm{IF} (X^\prime, \Sigma_{XX}^{-\frac{1}{2}}f_{jX})
= \Sigma_{XX}^{-\frac{1}{2}} \rm{IF} (X^\prime, f_{jX})+\rm{IF} (X^\prime, \Sigma_{XX}^{-\frac{1}{2}})f_{jX}
\label{CV1a}
\end{align}
By the first term of Eq. (\ref{CV1a}) we have
\begin{align}
&\Sigma_{XX}^{-\frac{1}{2}} \rm{IF} (X^\prime,  Y^\prime,f_{jX})= \Sigma_{XX}^{-\frac{1}{2}} ( \mb{B}-\rho^2\vc{I})^{-1}\rm{IF} (X^\prime, Y^\prime,\mb{B})f_{jX} \nonumber\\
&= - \Sigma_{XX}^{-\frac{1}{2}} ( \mb{B}-\rho^2\vc{I})^{-1}\big[ \Sigma_{XX}^{-\frac{1}{2}}\tilde{k}_X(\cdot, X^\prime)\otimes  \tilde{k}_X(\cdot, X^\prime)\Sigma_{XX}^{-\frac{1}{2}} \Sigma_{XY} \Sigma_{YY}^{-1}\Sigma_{XY}\Sigma_{XX}^{-\frac{1}{2}}\nonumber\\
&+ 2  \Sigma_{XX}^{-\frac{1}{2}}\Sigma_{XY}\Sigma^{-1}_{YY}\tilde{k}_X(\cdot, X^\prime)\otimes  \tilde{k}_Y(\cdot, Y^\prime) \Sigma_{XX}^{-\frac{1}{2}}-  \Sigma_{XX}^{-\frac{1}{2}}\Sigma_{XY}\Sigma_{YY}\tilde{k}_Y(\cdot, Y^\prime) \Sigma^{-1}_{YY}\Sigma_{YX}\Sigma_{XX}^{-\frac{1}{2}}\big]\tilde{f}_{jX}
\label{CV2a}
\end{align}
We derive each term of Eq. (\ref{CV2a}), respectively. The first term of
Eq. (\ref{CV2a}) is given by
\begin{align}
&\Sigma_{XX}^{-\frac{1}{2}} (\mb{B}-\rho^2\vc{I})^{-1}[ \Sigma_{XX}^{-\frac{1}{2}} \langle \tilde{k}_X(\cdot, X^\prime)\otimes \tilde{ k}_X(\cdot, X^\prime),  \Sigma_{XX}^{-\frac{1}{2}} \Sigma_{XY} \Sigma_{YY}^{-1}\Sigma_{XY}\Sigma_{XX}^{-\frac{1}{2}} f_{jX}\rangle \nonumber\\
&=\mb{L}_{jX}\langle\tilde{ k}_X(\cdot, X^\prime)\otimes \tilde{ k}_X(\cdot, X^\prime),  \Sigma_{XX}^{-\frac{1}{2}} \Sigma_{XY} \Sigma_{YY}^{-1}\Sigma_{YX} f_{jX}\rangle \nonumber\\
&= \mb{L}_{jX} \rho_j^2\langle\tilde{ k}_X(\cdot, X^\prime)\otimes \tilde{ k}_X(\cdot, \vc{x}^\prime),  f_{jX}\rangle  \nonumber\\
&= \mb{L}_{jX} \rho_j^2\langle\tilde{ k}_X(\cdot, X^\prime)\otimes \tilde{ k}_X(\cdot, X^\prime),  f_{jX}\rangle  \\ \nonumber
&= \mb{L}_{jX} \rho_j^2\tilde{f}(X^\prime) \tilde{k}(\cdot, X^\prime) \nonumber
\end{align}
The 2nd term of Eq. (\ref{CV2a}) is
\begin{align}
&2\mb{L}_{jX}  \Sigma_{XX}^{-\frac{1}{2}}\Sigma_{XY}\Sigma^{-1}_{YY}\tilde{k}_X(\cdot, X^\prime) \tilde{k}_Y(\cdot, Y^\prime) \Sigma_{XX}^{-\frac{1}{2}} f_{jX} \nonumber\\
&= \mb{L}_{jX} \langle  \Sigma_{XX}^{-\frac{1}{2}}\Sigma_{XY}\Sigma^{-1}_{YY} f_{jX},  \tilde{k}_Y(\cdot, Y^\prime)\rangle \tilde{k}_X(\cdot, X^\prime) + \mb{L}_{jX} \langle \Sigma_{XX}^{-\frac{1}{2}}\Sigma_{XY}\Sigma^{-1}_{YY}\tilde{k}_X(\cdot, X^\prime),  f_{jX}\rangle \tilde{k}_Y(\cdot, Y^\prime)\nonumber\\
&= \mb{L}_{jX}  \rho_j \tilde{f}_{jY}(Y^\prime)(\tilde{k}_X(\cdot, X^\prime)+ \mb{L}_{jX} \Sigma_{XX}^{-\frac{1}{2}}\Sigma_{XY}\Sigma^{-1}_{YY} \tilde{f}_{jX} (X^\prime) \tilde{k}_Y(\cdot, Y^\prime)   \nonumber
\end{align}
and the 3rd term of   Eq. (\ref{CV2a}) is
\begin{align}
&\mb{L}_{jX} \Sigma_{XY}\Sigma^{-1}_{YY}\tilde{k}_Y(\cdot, Y^\prime) \otimes \tilde{k}_Y(\cdot, Y^\prime) \Sigma_{YY}^{-1} \Sigma_{YX}\Sigma_{XX}^{-\frac{1}{2}}]f_{jX} \nonumber \\
&=\mb{L}_{jX} \langle  \Sigma_{XY}\Sigma^{-1}_{YY}\tilde{k}_Y(\cdot, Y^\prime) ,  \Sigma_{YY}^{-1}\Sigma_{YX}f_{jX}\rangle \tilde{k}_Y(\cdot, Y^\prime)\rangle \nonumber\\
&=\mb{L}_{jX} \langle  \Sigma_{XY}\Sigma^{-1}_{YY}\tilde{k}_Y(\cdot, Y^\prime) , \rho _jf_{jX}\rangle \tilde{k}_Y(\cdot, Y^\prime) \rangle\nonumber\\
&=\mb{L}_{jX} \rho _j   \Sigma_{XY}\Sigma^{-1}_{YY} \tilde{f}_{jY} (Y^\prime) \tilde{k}_Y(\cdot, Y^\prime) \nonumber
\end{align}
By substituting the above three equations into Eq. (\ref{CV2a}),  we have
\begin{align}
 &\Sigma_{XX}^{-\frac{1}{2}} \rm{IF} (\cdot, Z^\prime, f_{jX}) \nonumber\\&= \Sigma_{XX}^{-\frac{1}{2}} ( \mb{B} - \rho^2_j\vc{I})^{-1}\rm{IF} (\cdot, Z^\prime,\mb{B})f_{jX}
\nonumber\\&=-\rho_j (\tilde{f}_{jY}(Y^\prime)-\rho_j \tilde{f}_{jX}(X^\prime))\mb{L}_{jX} \tilde{k} (\cdot, X^\prime) - (\tilde{f}_{jX}(X^\prime) - \rho_j \tilde{f}_{jY}(Y^\prime)) \Sigma_{XY}\Sigma^{-1}_{YY} \tilde{k}_Y(\cdot, Y^\prime)
\label{CV6a}
\end{align}
The 2nd term of Eq. (\ref{CV1a}) is given 
\begin{align}
&\rm{IF} (X^\prime, \Sigma_{XX}^{-\frac{1}{2}}) f_{jX} \nonumber \\
&= - \langle f_{jX}, \Sigma_{XX}^{-1} f_{jX}\rangle  \Sigma_{XX}^{-\frac{1}{2}} \rm{IF} (X^\prime, \Sigma_{XX}^{\frac{1}{2}}) \Sigma_{XX}^{-\frac{1}{2}} e_{jX} \nonumber \\
&=  \langle f_{jX}, \Sigma_{XX}^{-\frac{1}{2}} \rm{IF} (X^\prime, \Sigma_{XX}^{\frac{1}{2}}) f_{jX}\rangle f_{jX}\nonumber \\
&= -\frac{1}{2}[\langle  f_{jX}, \Sigma_{XX}^{-\frac{1}{2}} \rm{IF} (X, \Sigma_{XX}^{\frac{1}{2}}) f_{jX}\rangle + \langle  f_{jX},  \rm{IF} (X^\prime, \Sigma_{XX}^{\frac{1}{2}}) \Sigma_{XX}^{-\frac{1}{2}} f_{jX}\rangle]f_{jX} \nonumber \\
&=  - \frac{1}{2}[\langle  f_{jX}, \Sigma_{XX}^{-\frac{1}{2}} \rm{IF} (X^\prime, \Sigma_{XX}) f_{jX}\rangle]f_{jX} \nonumber \\
&= -\frac{1}{2}[ \langle f_{jX}, (\tilde{k}_X(\cdot, X^\prime)-\Sigma_{XX} f_{jX})]f_{jX} \nonumber \\
&= -\frac{1}{2}[\tilde{f}_{jX}(X^\prime)- \langle f_{jX},\Sigma f_{jX}\rangle]f_{jX}\nonumber\\
&= \frac{1}{2}[1- \tilde{f}_{jX}(X^\prime)] f_{jX}
\label{CV7a}
\end{align}
Therefore, substituting  Eq. (\ref{CV6a})  and  Eq. (\ref{CV7a})   into  Eq. (\ref{CV1a}) we get the {\it {j}-th IF  of kernel CV for  the $X$ data:
\begin{multline}
 \rm{IF} (\cdot, X^\prime, Y^\prime,f_{jX})=-\rho_j (\tilde{f}_{jY}(Y^\prime)-\rho_j \tilde{f}_{jX}(X^\prime))\mb{L}_{jX} \tilde{k}_X(\cdot, X^\prime)-  (\tilde{f}_{jX}(X^\prime)-\rho_j \tilde{f}_{jY}(Y^\prime))\mb{L}_{jX} \Sigma_{XY}\Sigma^{-1}_{YY} \tilde{k}_Y(\cdot, Y^\prime)\\ +\frac{1}{2}[1- \tilde{f}^2(Y^\prime)]f_{jX} \nonumber
\end{multline}
Similarly, for the  $Y$ data we have,
\begin{multline}
 \rm{IF} (\cdot, X^\prime, Y^\prime,f_{jY})=-\rho_j (\tilde{f}_{jX}(X^\prime)-\rho_j \tilde{f}_{jY}(Y^\prime))\mb{L}_{jY} \tilde{k}_Y (\cdot, Y^\prime)-  (\tilde{f}_{jY}(Y^\prime)-\rho_j \tilde{f}_{jX}(X^\prime))\mb{L}_{jY} \Sigma_{YX}\Sigma^{-1}_{XX} \tilde{k}_X(\cdot, X^\prime)\\ +\frac{1}{2}[1- \tilde{f}^2(X^\prime)]f_{jY} \nonumber
\end{multline}

\section*{Appendix B:  Abbreviations  and symbols}
Table \ref{tb:lab} and  \ref{tb:lsy} preset the list of abbreviations, and  symbols, respectively. 
\begin{table}[h]
 \begin{center}
\caption {List of abbreviations }
\label{tb:lab}
\scalebox{0.8}{
 \begin{tabular}{llll} \hline
\rm{Abbreviation}&\rm{Elaboration} &\rm{Abbreviation}&\rm{Elaboration}\\ \hline
\rm{CC}&\rm{Canonical correlation}&\rm{CCA} &\rm{Canonical correlation Analysis}\\
\rm{CO}& \rm{Kernel covariance operator}&\rm{CCO}& \rm{Kernel cross-covariance operator}\\
\rm{CV}& \rm{Canonical Variates}& \rm{DE}&\rm{Density estimation} \\
\rm{DNA}&\rm{Deoxyribonucleic acid}&\rm{EIF} & \rm{Empirical influence function}\\
\rm{IF} & \rm{Influence function}&\rm{fMRI}& \rm{Functional magnetic resonance imaging}\\
\rm{IRWLS}&\rm{Iteratively re-weighted least squares} &\rm{KIRWLS}&\rm{Kerneled iteratively re-weighted least squares} \\
\rm{LUSC}&\rm{lung squamous cell carcinoma}&\rm{MCIC}& \rm{ The mind clinical imaging consortium}\\
\rm{ME}&\rm{Mean element} &\rm{MGSD} &\rm{ Multivariate Gaussian structural data}\\
\rm{NIH}&\rm{National institutes of health}&\rm{NSF}&\rm{National Science Foundation}\\
\rm{OVSC}& \rm{Ovarian serous cystadenocarcinoma}&\rm{PCA}& \rm{Principal component analysis}\\
\rm{PDK}&\rm{Positive definite kernel}&\rm{RKHS}&\rm{Reproducing kernel Hilbert Space}\\
\rm{SCFSD}&\rm{Sine cosine function  structural data}&\rm{SFSD}&\rm{Sine function  of  structural data}\\
\rm{SNP}&{Single-nucleotide polymorphism }&\rm{SZ}&\rm{Schizophrenia}\\
\rm{TCSD}&\rm{ Three circles structural data}&\rm{TCGA}&\rm{The cancer genome atlas}\\\hline
\end{tabular}
}
\end{center}
\end{table}

\begin{table}
 \begin{center}
\caption {List of  symbols.}
\label{tb:lsy}
\scalebox{0.65}{
 \begin{tabular}{llll} \hline
\rm{Symbol}&\rm{Explanation} & \rm{Symbol}&\rm{Explanation}\\  \hline
$\mb{R}$& \rm{The set of real numbers}& $\Omega$& \rm{  A sample space}\\
$\mc{A}$& \rm{A set of events}& $P$& \rm{ A function  from events to probabilities}\\
$\vc{\Phi}(\cdot)$& \rm{Feature map}& $\tilde{\vc{\Phi}}(\cdot)$& \rm{Centered feature map}\\
$\mc{H}_X$& \rm{RKHS of X data}& $\mc{H}_X$& \rm{RKHS of Y data}\\
$\mc{H}$& \rm{Hilbert space}& $\mc{B}_\mc{H}$& \rm{$\sigma$-filed of Borel sets }\\
$k_X(X_i, X_j)$& \rm{PDK of X data }&$k_Y(Y_i, Y_j)$& \rm{PDK of Y data }\\
$\tilde{k}_X(X_i, X_j)$& \rm{ Centered PDK of X data }&$\tilde{k}_Y(Y_i, Y_j)$& \rm{ Centered PDK of Y data }\\
$k_X(\cdot, X)$& \rm{$\mc{H}_X$-valued random variable}&$k_Y(\cdot, Y)$& \rm{$\mc{H}_Y$-valued random variable}\\
$F_X$& \rm{Probability distribution  of  X }& $F_Y$& \rm{ Probability distribution of Y}\\
$F_{XY}$& \rm{Joint probability distribution  of  (X,Y) }& $F_{nXY}$& \rm{ Empirical joint probability distribution.}\\
$E_X$& \rm{Expectation   of  X }& $E_Y$& \rm{ Expectation  of Y}\\
$\mc{M}_X$& \rm{ Kernel mean element  of  X }& $\mc{M}_Y$& \rm{ Kernel mean element  of  Y}\\
$\widehat{\mc{M}}_X$& \rm{Estimated kernel mean element  of  X }& 
$\widehat{\mc{M}}_R$& \rm{Robust kernel mean element}\\
 $\zeta(\cdot)$& \rm{ Robust loss function}&  $\varphi(t)$&  $\frac{\zeta^\prime(t)}{t}$, Weight function\\
$\vc{K}_X$& \rm{Gram matrix  of  X data}& $\vc{K}_Y$& \rm{ Gram matrix of Y data}\\
$\vc{G}_X$& \rm{Centered Gram matrix  of  X data}& $\vc{G}_Y$& \rm{  Centered Gram matrix of Y data}\\
$\rho$& \rm{Robust kernel canonical  correlation}& $\hat{\rho}$& \rm{Estimate  robust kernel canonical correlation}\\
$\vc{a}_{X}$& \rm{Canonical direction  of  X data}& $\vc{a}_{Y}$& \rm{Canonical direction of Y data}\\
$\vc{G}_{RX}$& \rm{ Robust centered Gram matrix  of  X data}& $\vc{G}_{RY}$& \rm{  Robust centered Gram matrix of Y data}\\
$\vc{b}_{X}$& \rm{ Robust  canonical direction  of  X data}& $\vc{b}_{Y}$& \rm{  Robust canonical direction of Y data}\\
$\rho_{rkcc} $& \rm{Kernel canonical  correlation}& $\hat{\rho}_{rkcc}$& \rm{Estimate  kernel canonical correlation}\\
$\Sigma _{XX}$& Kernel CO& $\Sigma _{XY}$& \rm{ Kernel CCO}\\
$\Sigma _{RXX}$& Robust kernel CO& $\Sigma _{RXY}$& \rm{Robust kernel CCO}\\
$\widehat{\Sigma} _{XX}$& \rm{Estimate of the  kernel CO}& $\widehat{\Sigma} _{XY}$& \rm{Estimate of the  kernel CCO}\\
$\widehat{\Sigma} _{RXX}$& \rm{ Estimate of the robust kernel CO}& $\widehat{\Sigma} _{RXY}$& \rm{Estimate of the  robust kernel CCO}\\
$f_{X}$& \rm{Canonical projection along eigen-function of  X data}& $f_{Y}$& \rm{ Canonical projection along eigen-function of Y data}\\
$g_{X}$& \rm{Robust canonical projection along eigen-function of  X data}& $g_{Y}$& \rm{ Robust canonical projection along eigen-function of Y data}\\
$\eta_{KCO}$& \rm{ Performance measure on the kernel CO}& $\eta_{RKCO}$& \rm{ Robust performance measure of the kernel CO}\\
$\eta_{\rho}$& \rm{Performance measure on the kernel CC}& $\eta_{f}$& \rm{ Performance measure of the kernel CV}\\\hline
\end{tabular}
}
\end{center}
\end{table}
\newpage
\section*{References}
\bibliography{Ref-UKIF}

\begin{thebibliography}{10}
\expandafter\ifx\csname url\endcsname\relax
  \def\url#1{\texttt{#1}}\fi
\expandafter\ifx\csname urlprefix\endcsname\relax\def\urlprefix{URL }\fi
\expandafter\ifx\csname href\endcsname\relax
  \def\href#1#2{#2} \def\path#1{#1}\fi

\bibitem{SVM92}
B.~E. Boser, I.~M. Guyon, V.~N. Vapnik, A training algorithm for optimal margin
  classifiers, in: D.~Haussler (Ed.), Fifth Annual ACM Workshop on
  Computational Learning Theory, ACM Press, Pittsburgh, PA, 1992, pp. 144--152.

\bibitem{Saunders98}
C.~Saunders, A.~Gammerman, V.~Vovk, Ridge regression learning algorithm in dual
  variables, in: Proceedings of the 15th International Conference on Machine
  Learning (ICML1998), Morgan Kaufmann, San Francisco, CA, 1998, pp. 515--521.

\bibitem{Charpiat-15}
G.~Charpiat, M.~Hofmann, B.~Sch{\"o}lkopf, Kernel methods in medical imaging,
  Springer, Berlin, Germany, Ch.~4, pp. 63--81.

\bibitem{Back-08}
F.~R. Bach, Consistency of the group lasso and multiple kernel learning,
  Journal of Machine Learning Research 9 (2008) 1179--1225.

\bibitem{Steinwart-08}
I.~Steinwart, A.~Christmann, Support Vector Machines, Springer, New York, 2008.

\bibitem{Hofmann-08}
T.~Hofmann, B.~Sch{\"{o}}lkopf, J.~A. Smola, Kernel methods in machine
  learning, The Annals of Statistics 36 (2008) 1171--1220.

\bibitem{Schlkof-kpca}
B.~Sch{\"{o}}lkopf, A.~J. Smola, K.-R. M{\"{u}}ller, Nonlinear component
  analysis as a kernel eigenvalue problem, Neural Computation. 10 (1998)
  1299--1319.

\bibitem{Akaho}
S.~Akaho, A kernel method for canonical correlation analysis, International
  meeting of psychometric Society. 35 (2001) 321--377.

\bibitem{Back-02}
F.~R. Bach, M.~I. Jordan, Kernel independent component analysis, Journal of
  Machine Learning Research 3 (2002) 1--48.

\bibitem{Ashad-14}
M.~A. Alam, K.~Fukumizu, Hyperparameter selection in kernel principal component
  analysis, Journal of Computer Science 10(7) (2014) 1139--1150.

\bibitem{Yu-11}
S.~Yu, L.-C. Tranchevent, B.~D. Moor, Y.~Moreau, Kernel-based Data Fusion for
  Machine Learning, Springer, Verlag Berlin Heidelberg, 2011.

\bibitem{Christmann-04}
A.~Christmann, I.~Steinwart, On robustness properties of convex risk
  minimization methods for pattern recognition, Journal of Machine Learning
  Research 5 (2004) 1007--1034.

\bibitem{Christmann-07}
A.~Christmann, I.~Steinwart, Consistency and robustness of kernel-based
  regression in convex risk minimization, Bernoulli 13(3) (2007) 799--819.

\bibitem{Debruyne-08}
M.~Debruyne, M.~Hubert, J.~Horebeek, Model selection in kernel based regression
  using the influence function, Journal of Machine Learning Research 9 (2008)
  2377--2400.

\bibitem{Huber-09}
P.~J. Huber, E.~M. Ronchetti, Robust Statistics, John Wiley \& Sons, England,
  2009.

\bibitem{Hampel-11}
F.~R. Hampel, E.~M. Ronchetti, P.~J. Rousseeuw, W.~A. Stahel, Robust
  Statistics: The Approach Based on Influence Functions, John Wiley \& Sons,
  New York, 2011.

\bibitem{Kim-12}
J.~Kim, C.~D. Scott, Robust kernel density estimation, Journal of Machine
  Learning Research 13 (2012) 2529--2565.

\bibitem{Huang-KPCA}
S.~Y. Huang, Y.~R. Yeh, S.~Eguchi, Robust kernel principal component analysis,
  Neural Computation 21(11) (2009) 3179--3213.

\bibitem{Debruyne-10}
M.~Debruyne, M.~Hubert, J.~Horebeek, Detecting influential observations in
  kernel {\mbox{pca}}, Computational Statistics and Data Analysis 54 (2010)
  3007--3019.

\bibitem{Fukumizu-SCKCCA}
K.~Fukumizu, F.~R. Bach, A.~Gretton, Statistical consistency of kernel
  canonical correlation analysis, Journal of Machine Learning Research 8 (2007)
  361--383.

\bibitem{Hardoon2009}
D.~R. Hardoon, J.~Shawe-Taylor, Convergence analysis of kernel canonical
  correlation analysis: theory and practice, Machine Learning 74 (2009) 23--38.

\bibitem{Otopal-12}
N.~Otopal, Restricted kernel canonical correlation analysis, Linear Algebra and
  its Applications 437 (2012) 1--13.

\bibitem{Ashad-15}
M.~A. Alam, K.~Fukumizu, Higher-order regularized kernel canonical correlation
  analysis, International Journal of Pattern Recognition and Artificial
  Intelligence 29(4) (2015) 1551005(1--24).

\bibitem{Ashad-10}
M.~A. Alam, M.~Nasser, K.~Fukumizu, A comparative study of kernel and robust
  canonical correlation analysis, Journal of Multimedia. 5 (2010) 3--11.

\bibitem{Romanazii-92}
M.~Romanazzi, Influence in canonical correlation analysis, Psychometrika 57(2)
  (1992) 237--259.

\bibitem{Gretton-08}
A.~Gretton, K.~Fukumizu, C.~H. Teo, L.~Song, B.~Sch{\"{o}}lkopf, A.~Smola, A
  kernel statistical test of independence, In Advances in Neural Information
  Processing Systems 20 (2008) 585--592.

\bibitem{Fukumizu-08}
K.~Fukumizu, A.~Gretton, X.~Sun, B.~Sch{\"{o}}lkopf, Kernel measures of
  conditional dependence, In Advances in Neural Information Processing Systems,
  Cambridge, MA, MIT Press 20 (2008) 489�496.

\bibitem{Song-08}
L.~Song, A.~Smola, K.~Borgwardt, A.~Gretton, Colored maximum variance
  unfolding, Advances in Neural Information Processing Systems 20 (2008)
  1385--1392.

\bibitem{Gretton-12}
A.~Gretton, K.~M. Borgwardt, M.~J. Rasch, B.~Sch{\"{o}}lkopf, A.~J. Smola, A
  kernel two-sample test, Journal of Machine Learning Research 13 (2012) 723 --
  773.

\bibitem{Aron-RKHS}
N.~Aronszajn, Theory of reproducing kernels, Transactions of the American
  Mathematical Society 68 (1950) 337--404.

\bibitem{Berlinet-04}
A.~Berlinet, C.~Thomas-Agnan, Reproducing kernel Hilbert spaces in probability
  and statistics, Kluwer Academic Publishers, London, 2004.

\bibitem{Ashad-14T}
M.~A. Alam, Kernel Choice for Unsupervised Kernel Methods, PhD. Dissertation,
  The Graduate University for Advanced Studies, Japan, 2014.

\bibitem{Fukumizu-14}
K.~Fukumizu, C.~Leng, Gradient-based kernel dimension reduction for regression,
  Journal of the American Statistical Association 109(550) (2014) 359--370.

\bibitem{Hampel-86}
F.~R. Hampel, E.~M. Ronchetti, W.~A. Stahel, Robust Statistics, John Wiley \&
  Sons, New York, 1986.

\bibitem{Tukey-77}
J.~W. Tukey, Exploratory Data Analysis, Addison-Wesley, Reading, Massachusetts,
  1977.

\bibitem{Reed-80}
M.~Reed, B.~Simon, Methods of Modern Mathematical Physics, Academic Press,
  California, 1980.

\bibitem{Luenberger-97}
D.~G. Luenberger, Optimization by Vector Space Methods, Wiley-Interscience, New
  York, 1997.

\bibitem{Drineas-05}
P.~Drineas, M.~W. Mahoney, On the nystr{\"{o}}m method for approximating a gram
  matrix for improved kernel-based learning, Journal of Machine Learning
  Research 6 (2005) 2153--2175.

\bibitem{Rahimi-07}
A.~Rahimi, B.~Recht, Random features for large-scale kernel machines, In Neural
  Information Processing Systems (NIPS) 3 (2007) 5.

\bibitem{Huber-64}
P.~J. Huber, Robust estimation of a location parameter, Annals of Mathematical
  Statistics 35 (1964) 73--101.

\bibitem{Hampel-74}
F.~R. Hampel, The influence curve and its role in robust estimations, Journal
  of the American Statistical Association 69 (1974) 386--393.

\bibitem{Dono-83}
D.~L. Donoho, P.~J. Huber, The notion of breakdown point, In P. J. Bickel, K.
  A. Doksum, and J. L. Hodges Jr, editors, {\em A Festschrift for Erich L.
  Lehmann}, Belmont, California, Wadsworth 12(3) (1983) 157--184.

\bibitem{Lai-00}
P.~Lai, C.~Fyfe, Kernel and nonlinear canonical correlation analysis, Computing
  and Information Systems 7 (2000) 43--49.

\bibitem{Alzate2008}
C.~Alzate, J.~A.~K. Suykens, A regularized kernel {{CCA}} contrast function for
  {{ICA}}, Neural Networks 21 (2008) 170--181.

\bibitem{Hardoon2004}
D.~R. Hardoon, S.~Szedmak, J.~Shawe-Taylor, Canonical correlation analysis: an
  overview with application to learning methods, Neural Computation 16 (2004)
  2639--2664.

\bibitem{Huang-2009}
S.~Y. Huang, M.~Lee, C.~Hsiao, Nonlinear measures of association with kernel
  canonical correlation analysis and applications, Journal of Statistical
  Planning and Inference 139 (2009) 2162--2174.

\bibitem{Anderson-03}
T.~W. Anderson, An Introduction to Multivariate Statistical Analysis, John
  Wiley\& Sons, third edition, 2003.

\bibitem{Tanaka-88}
Y.~Tanaka, Sensitivity analysis in principal component analysis: influence on
  the subspace spanned by principal components, Communications in
  Statistics-Theory and Methods 17(9) (1988) 3157--3175.

\bibitem{Tanaka-89}
Y.~Tanaka, Influence functions related to eigenvalue problem which appear in
  multivariate analysis, Communications in Statistics-Theory and Methods 18(11)
  (1989) 3991--4010.

\bibitem{Adrover-15}
J.~Adrover, S.~M. Donato, A robust predictive approach for canonical
  correlation analysis, Journal of Multivariate Analysis. 133 (2015) 356--376.

\bibitem{Schlkof-book}
B.~Sch{\"{o}}lkopf, A.~J. Smola, Learning with Kernels, MIT Press, Cambridge
  MA, 2002.

\bibitem{Parkhomenko-09}
E.~Parkhomenko, D.~Tritchler, J.~Beyene, Sparse canonical correlation analysis
  with application to genomic data integration, Statistical Applications in
  Genetics and Molecular Biolog 8(1) (2009) 1--34.

\bibitem{Sequeira-11}
J.~Sequeira, A.~Tsourdos, S.~B. Lazarus, Robust covariance estimation for data
  fusion from multiple sensors, IEEE Transactions on Instrumentation and
  Measurement 60(12) (2011) 3833--13844.

\bibitem{Ashad-16a}
M.~A. {Alam}, Y.-P. {Wang}, {Influence Function of Multiple Kennel Canonical
  Analysis to Identify Outlier in Imaging Genetics Data}, ArXiv e-prints\href
  {http://arxiv.org/abs/1606.00113} {\path{arXiv:1606.00113}}.

\end{thebibliography}

\end{document}